%% file: acl_latex.tex
\pdfoutput=1

\documentclass[11pt]{article}

\usepackage[]{acl}

\usepackage{times}
\usepackage{latexsym}
\usepackage{amsmath}
\usepackage{graphicx} 
\usepackage{pifont}
\usepackage{listings} 

\usepackage[T1]{fontenc}

\usepackage[utf8]{inputenc}

\usepackage{microtype}

\usepackage{inconsolata}

\title{Selective Reflection-Tuning:\\Student-Selected Data Recycling for LLM Instruction-Tuning}

\author{
    Ming Li\textsuperscript{\rm 1}, Lichang Chen\textsuperscript{\rm 1}, Jiuhai Chen\textsuperscript{\rm 1}, Shwai He\textsuperscript{\rm 1}, Jiuxiang Gu\textsuperscript{\rm 2}, Tianyi Zhou\textsuperscript{\rm 1}\\
    \textsuperscript{\rm 1}University of Maryland, College Park~~~~
    \textsuperscript{\rm 2}Adobe Research\\
    \texttt{\{minglii, bobchen, tianyi\}@umd.edu} \\
    Project: \url{https://github.com/tianyi-lab/Reflection_Tuning}
}

\begin{document}
\maketitle
\begin{abstract}
Instruction tuning is critical to large language models (LLMs) for achieving better instruction following and task adaptation capabilities but its success heavily relies on the training data quality. Many recent methods focus on improving the data quality but often overlook the compatibility of the data with the student model being finetuned. This paper introduces Selective Reflection-Tuning, a novel paradigm that synergizes a teacher LLM's reflection and introspection for improving existing data quality with the data selection capability of the student LLM, to automatically refine existing instruction-tuning data. This teacher-student collaboration produces high-quality and student-compatible instruction-response pairs, resulting in sample-efficient instruction tuning and LLMs of superior performance. Selective Reflection-Tuning is a data augmentation and synthesis that generally improves LLM finetuning and self-improvement without collecting brand-new data. We apply our method to Alpaca and WizardLM data and achieve much stronger and top-tier 7B and 13B LLMs. 
\end{abstract}

\input{introduction}

\input{method}

\begin{table*}[h]
\centering
\scalebox{0.75}{
\begin{tabular}{l|cccccccc}
\hline
& \multicolumn{7}{c}{\textbf{Alpaca Eval Leaderboard}}  \\
 & \textbf{Win Rate} & \textbf{Standard Error} & \textbf{Wins} & \textbf{Draws} & \textbf{Avg Length} & \textbf{Data} & \textbf{RLHF/AIF} \\
\hline
GPT4 \cite{openai2023gpt4} & 95.28 & 0.72 & 761 & 12 & 1365 & / & /\\
Claude 2 & 91.36 & 0.99 & 734 & 1 & 1069 & / & / \\
Zephyr 7B Beta \cite{tunstall2023zephyr} & 90.60 & 1.03 & 727 & 1 & 1444 & 774,000 & \ding{51} \\
ChatGPT & 89.37 & 1.08 & 716 & 5 & 827 & / & / \\
Evo v2 7B  & 89.35 & 1.08 & 715 & 5 & 1754 & / & / \\
XwinLM 7b V0.1 \cite{xwin-lm} & 87.83 & 1.15 & 703 & 1 & 1894 & / & \ding{51} \\
\textbf{sRecycled WizardLM 13B (ours)} & \textbf{85.96} & \textbf{1.23} & \textbf{692} & \textbf{0} & \textbf{1523} & \textbf{46,064} & \ding{55}\\
Zephyr 7B Alpha \cite{tunstall2023zephyr} & 85.76 & 1.23 & 688 & 3 & 1302 & 774,000 & \ding{51} \\
OpenChat V2 13B \cite{wang2023openchat} & 84.97 & 1.26 & 683 & 2 & 1564 & 82,600 & \ding{55} \\
Humpback LLaMa 65B \cite{li2023self} & 83.71 & 1.31 & 672 & 2 & 1269 & 502,133 & \ding{55} \\
UltraLM 13B V2.0 \cite{ding2023enhancing} & 80.64 & 1.31 & 673 & 0 & 1399 & 774,000 & \ding{55} \\
\textbf{sRecycled WizardLM 7B (ours)} & \textbf{83.48} & \textbf{1.31} & \textbf{672} & \textbf{0} & \textbf{1583} & \textbf{46,325} & \ding{55}\\
Vicuna 13B v1.3 \cite{vicuna2023} & 82.11 & 1.35 & 660 & 2 & 1132 & 125,000 & \ding{55} \\
GPT-3.5  & 81.71 & 1.33 & 642 & 25 & 1018 & / & / \\
LLaMA2 Chat 13B \cite{touvron2023llama2} & 81.09 & 1.38 & 652 & 0 & 1513 & 27,750 & \ding{51} \\
UltraLM 13B \cite{ding2023enhancing} & 80.64 & 1.40 & 647 & 1 & 1087 & 774,000 & \ding{55} \\
\textbf{sRecycled Alpaca 7B (ours)} & \textbf{79.58} & \textbf{1.42} & \textbf{639} & \textbf{0} & \textbf{1353} & \textbf{37,114} & \ding{55}\\
Claude2 Alpaca 13B \cite{claude2-alpaca} & 78.93 & 1.44 & 633 & 0 & 1127 & 52,002 & \ding{55} \\
Recycled WizardLM 7B  & 78.88 & 1.44 & 635 & 0 & 1494 & 70,000 & \ding{55} \\
Recycled Alpaca 7B  & 76.99 & 1.49 & 619 & 0 & 1397 & 52,002 & \ding{55} \\
Vicuna 7B v1.3 \cite{vicuna2023} & 76.84 & 1.49 & 614 & 3 & 1110 & 125,000 & \ding{55} \\
WizardLM 13B \cite{xu2023wizardlm}& 75.31 & 1.51 & 601 & 9 & 985 & 250,000 & \ding{55} \\
Guanaco 65B \cite{dettmers2023qlora}& 71.80 & 1.59 & 578 & 0 & 1249 & 9,850 & \ding{55} \\
LLaMA2 Chat 7B \cite{touvron2023llama2}& 71.37 & 1.59 & 574 & 1 & 1479 & 27,750 & \ding{51} \\
Vicuna 7B \cite{vicuna2023} & 64.41 & 1.69 & 517 & 3 & 1044 & 70,000 & \ding{55} \\
Davinci003 & 50.00 & 0.00 & 0 & 805 & 307 & / & / \\
LIMA 7B \cite{zhou2023lima} & 41.29 & 1.74 & 332 & 0 & 1624 & 1,000 & \ding{55} \\
Alpaca 7B \cite{alpaca} & 26.46 & 1.54 & 205 & 16 & 396 & 52,002 & \ding{55} \\
\hline
\end{tabular}}
\caption{The comparison of performance on AlpacaEval Leaderboard. ``Data'' represents the number of data used for fine-tuning. ``RLHF/AIF'' represents whether the model utilize an additional RLHF or RLAIF process. }
\label{tbl:alpaca_eval}
\end{table*}

\begin{table*}[h]
\centering
\scalebox{0.8}{
\begin{tabular}{l|ccccccc}
\hline
& \multicolumn{7}{c}{\textbf{Huggingface Open LLM Leaderboard}}  \\
& \textbf{Average} & \textbf{ARC} & \textbf{HellaSwag} & \textbf{MMLU} & \textbf{TruthfulQA} & \textbf{Data} & \textbf{RLHF/AIF} \\
\hline
Alpaca 7B \cite{alpaca} & 50.21 & 42.65 & 76.91 & 41.73 & 39.55 & 52,002 & \ding{55} \\
WizardLM 7B \cite{xu2023wizardlm} & 54.18 & 51.60 & 77.70 & 42.70 & 44.70 & 70,000 & \ding{55} \\
Vicuna 7B v1.3 \cite{vicuna2023} & 55.63 & 50.43 & 76.92 & 48.14 & 47.01 & 125,000 & \ding{55} \\
\textbf{sRecycled Alpaca 7B (ours)} & 56.05 & 54.01 & 78.07 & 46.69 & 45.41 & 37,114 & \ding{55} \\
LLaMA2 Chat 7B \cite{touvron2023llama2}& 56.34 & 52.90 & 78.55 & 48.32 & 45.57 & 27,750 & \ding{51} \\
\textbf{sRecycled WizardLM 7B (ours)} & 56.79 & 54.78 & 77.86 & 45.63 & 48.91 & 46,325 & \ding{55} \\ 
Vicuna 13B v1.1 \cite{vicuna2023} & 59.21 & 52.73 & 80.14 & 51.90 & 52.08 & 125,000 & \ding{55} \\
LLaMA2 Chat 13B \cite{touvron2023llama2}& 59.94 & 59.04 & 81.94 & 54.64 & 44.12 & 27,750 & \ding{51} \\
Vicuna 13B v1.3 \cite{vicuna2023} & 60.01 & 54.61 & 80.41 & 52.88 & 52.14 & 125,000 & \ding{55} \\
\textbf{sRecycled WizardLM 13B (ours)} & 60.22 & 59.73 & 80.15 & 55.64 & 45.37 & 46,064 & \ding{55} \\ 
WizardLM 13B 1.0 \cite{xu2023wizardlm} & 60.25 & 57.20 & 81.00 & 52.30 & 50.50 & 250,000 & \ding{55} \\
\hline
\end{tabular}
}
\caption{
The comparison of performance on Huggingface Open LLM Leaderboard. ``Data'' represents the number of data used for fine-tuning. ``RLHF/AIF'' represents whether the model utilizes an additional RLHF or RLAIF process.
}
\label{tbl:open}
\end{table*}

\section{Experimental Setup}

\subsection{Base Datasets}

The Alpaca dataset \cite{alpaca}, sourced from Stanford University, offers $52,002$ instruction samples. Developed via the self-instruct paradigm \cite{wang-etal-2023-self-instruct}, it leveraged the capabilities of the text-davinci-003 model. 
The WizardLM dataset \cite{xu2023wizardlm} is a refined collection encompassing a total of $250,000$ instruction samples. To enhance data fidelity, gpt-3.5-turbo-0613 has been meticulously integrated during the refinement process. From this extensive dataset, we predominantly focused on the WizardLM-7b subset, comprising $70,000$ samples.
We test our method on both of these two datasets to verify the effectiveness of our method and name the corresponding models as ``sRecycled Alpaca'' and ``sRecycled WizardLM''. 
\looseness-1

\subsection{Evaluation Metric}

To evaluate the effectiveness of our method, we utilize 4 commonly used automatic evaluation metrics, including (1) \textbf{Pair-wise Comparison}, (2) \textbf{Alpaca Eval}, (3) \textbf{Open LLM Leaderboard}, and (4) \textbf{MT-Bench}. Besides, additional (5) \textbf{Human Study} is also conveyed for the evaluation. \footnote{Detailed description can be found in Appendix \ref{evaluation}.}

\begin{figure}[thb]
\centering 
\includegraphics[width=0.5\textwidth]{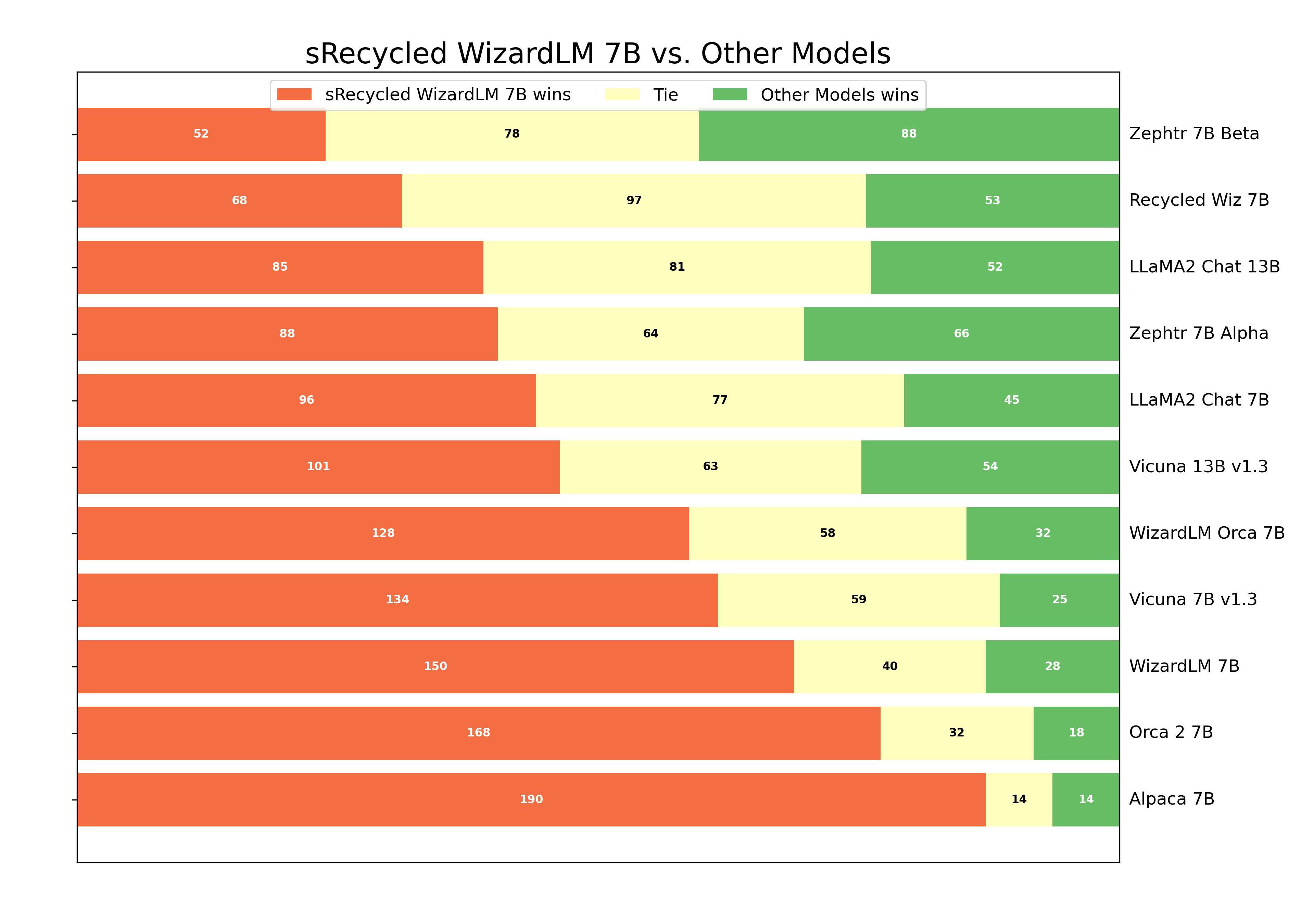} 
\caption{
The pair-wise comparison between our model with other classic open-source models by using GPT4 as the judge. From the comparison, our model outperforms most of them by a large margin, regardless of their model size and whether extra RLHF/AIF is utilized. 
} 
\label{pair_compare} 
\end{figure}

\begin{figure*}[!htbh]
\centering 
\includegraphics[width=1\textwidth]{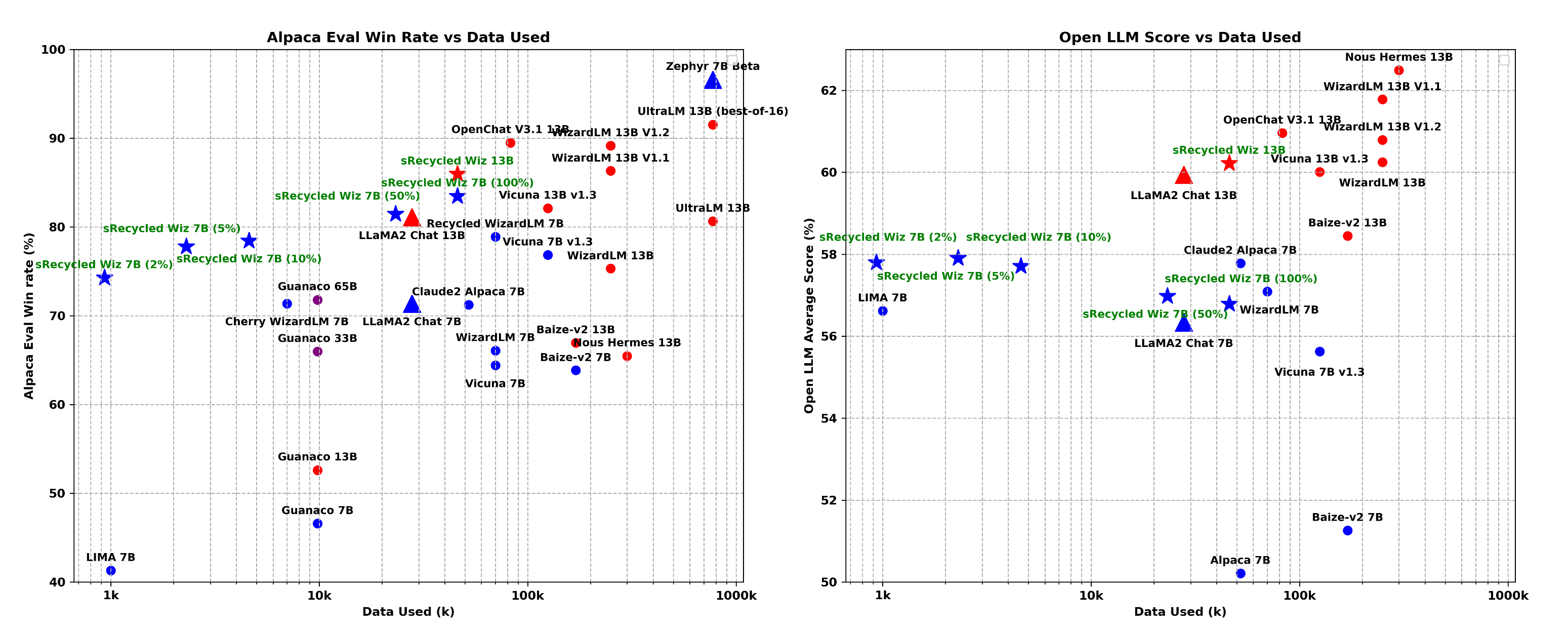} 
\caption{
Comparison between model performances and data used for fine-tuning on the Alapca Eval benchmark and the open LLM leaderboard. We utilize star markers to represent our models, dot markers to represent other instruction-tuned models and triangle markers to represent RLHF/AIF models. Blue markers represent 7B models, red markers represent 13B models and purple markers represent models with larger weights. 
} 
\label{ratio_plot} 
\end{figure*}

\begin{table*}[!htbh]
\centering
\scalebox{0.85}{
\begin{tabular}{l|ccccc|c}
\hline
& \multicolumn{5}{c|}{\textbf{Huggingface Open LLM Leaderboard}} & \textbf{AlpacaEval} \\
& \textbf{Average} & \textbf{ARC} & \textbf{HellaSwag} & \textbf{MMLU} & \textbf{TruthfulQA} & \textbf{AlpacaEval} \\
\hline
{sRecycled WizardLM 7B (2\%) (926)} & 57.80 & 54.69 & 78.80 & 47.00 & 50.70 & 74.29  \\
{sRecycled WizardLM 7B (5\%) (2,316)} & 57.91 & 54.86 & 79.83 & 46.69 & 50.23 & 77.78  \\
{sRecycled WizardLM 7B (10\%) (4,632)} & 57.71 & 55.46 & 79.56 & 46.83 & 48.98 & 78.43  \\
{sRecycled WizardLM 7B (30\%) (13,897)} & 56.89 & 54.61 & 79.25 & 44.67 & 49.05 & 82.48  \\
\hline
{sRecycled WizardLM 7B (100\%) (46,325)} & 56.79 & 54.78 & 77.86 & 45.63 & 48.91 & 83.21 \\
\hline
\end{tabular}
}
\caption{
The comparison of performance on Huggingface Open LLM Leaderboard and AlpacaEval Leaderboard by using different amounts of selective recycled WizardLM data. In the first parentheses are the percentage of data used for tuning and in the second parentheses are the specific amount of number used.  
}
\label{tbl:bench_few_few}
\end{table*}

\section{Experimental Results}

\subsection{Main Results}

For \textbf{Pair-wise Comparison}, we compare our sRecycled WizardLM 7B with other classic open-source models by using GPT4 as the judge as shown in Figure \ref{pair_compare}. 
Notably, our model outperforms most models by a large margin, regardless of whether they are 7B or 13B, (``LLaMA2 Chat 13B'', ``Vicuna 13B v1.3''), or whether extra RLHF/AIF is utilized (``LLaMA2 Chat 7B'', ``Zephyr 7B Alpha''), or whether other data improvement methods are utilized (``Recycled Wiz 7B'', ``WizardLM Orca 7B''\footnote{\url{https://huggingface.co/datasets/pankajmathur/WizardLM_Orca}}, ``Orca 2 7B''\cite{mitra2023orca}).

Table \ref{tbl:alpaca_eval} delineates the outcomes on the \textbf{AlpacaEval Leaderboard} in which our models stand out for delivering promising results with a streamlined approach. 
This comparison provides a direct quantification of a model's capacity for instruction adherence and the intrinsic quality of its output. Remarkably, with a win rate that competes closely with heavyweight counterparts, our models achieve this with only instruction tuning on a small amount of our high-quality data. Furthermore, our approach does not rely on additional processes such as RLHF \cite{NEURIPS2022_b1efde53, bai2022training} or RLAIF \cite{bai2022constitutional, lee2023rlaif}, which demand a significant overhead. This reduction in complexity represents a significant advancement in model efficiency, making it a cost-effective and agile solution for real-world applications. The ingenuity of our model lies in its simplicity and effectiveness, proving that with intelligent design less is more.


Table \ref{tbl:open} showcases the performance comparison on the \textbf{Huggingface Open LLM Leaderboard} with some related models. Similarly, with only instruction tuning on a small amount of data, our models surpass plenty of the models on the average performances across representative benchmarks. 
These benchmarks do not directly measure the instruction-following ability or the quality of responses generated by LLMs, but a relatively higher performance on these benchmarks still shows the non-degradation quality of our method.  


For the \textbf{human evaluation}, we compare the responses to given testing instructions between our sRecycled WizardLM 7B model with the original WizardLM 7B model by human evaluators, there are $57/108$ wins for our model, $23/108$ ties, and $28/108$ losses. These results further prove the efficacy of our method in improving the quality of the original data. \looseness-1

\subsection{Fewer Data Scenario}

To better illustrate the supreme quality of our sRecycled dataset, we further conduct experiments where only part of the data samples are utilized. Following \citet{li2023quantity}, we calculate the IFD score of each data sample and select the top $k$-percent of the data for the instruction tuning. Their performances on the Open LLM Leaderboard and the Alpaca Eval Leaderboard are shown in Table \ref{tbl:bench_few_few} \footnote{Detailed table and ablation can be found in Appendix \ref{few}}. Since selecting data by IFD score is an effective method to find a better instruction tuning subset from the overall data set, this consistent decrease in performance on Alpaca Eval indicates the difficulty in finding a subset with higher performances, which further verifies the overall high quality of our selective recycled data.

Figure \ref{ratio_plot} draws the scatters comparing the data used and corresponding performance. 
It illustrates a striking balance of efficiency and performance achieved by our models. 
Despite using markedly less data, our models—represented by the distinctive star markers—consistently occupy the upper echelons of the performance spectrum on both the Alpaca Eval benchmark and the open LLM leaderboard. 
Furthermore, the plots reveal that our models achieve these results without scaling up to the larger data requirements that other models seem to necessitate, as indicated by their position further to the right along the x-axis. 
The results not only signal superior data quality but also suggest a potential reduction in the computational resources and time required for training, which is crucial for sustainable and scalable AI development. 

Furthermore, it is astonishing that with less than $1,000$ selective recycled data, our ``sRecycled WizardLM 7B (2\%) (926)'' outperforms most existing 7B models, including LIMA, which is trained with manually curated data samples. 
This not only verifies LIMA's \cite{zhou2023lima} hypothesis but also pushes it further forward: In addition to human-carefully-crafted instruction tuning data, less than $1,000$ totally automatically generated data can also yield substantial benefits in model alignment and performance. \looseness-1

\section{Ablation Study}

\subsection{Ablation on Reflection}

Extensive experiments are conducted on several 7B models as shown in Table \ref{tbl:ablation}. We utilize the pair-wise comparison with GPT4 as the judge to measure the performance of different models.

Compared with the original WizardLM model, our performance is dramatically better, which directly showcases the supreme capability of our method to increase the data quality. ``Reflect on Ins.'' and ``Reflect on Res.'' represent models that are trained with data reflected merely on instruction or response and no selection process is utilized. Through these comparisons, it can be found that reflection on instruction only improves the data quality a little, while reflection on response improves the data quality more. This phenomenon is reasonable due to the similarity in response distribution between original WizardLM data and WizardLM data reflected on instruction. On the contrary, when the response is reflected, it directly affects the target that LLM needs to fit on, thus directly showing an improvement in the response quality. ``Reflect on Ins. + Res.'' represents the model trained by using reflection-tuning (``Recycled WizardLM 7B'') without the selection process, though already having the good capability to follow instructions, our model still outperforms it with less data. 

\begin{table}[!tbh]
\centering
\scalebox{0.8}{
\begin{tabular}{lcccc}
\hline
& Win & Tie & Lose & Win Rate\\\hline
vs. Original WizardLM & 150     & 40       & 28 & 1.560  \\
\hline
vs. Reflect on Ins.   & 143     & 51       & 24  & 1.546  \\
vs. Reflect on Res.   & 72     & 93       & 53  & 1.087  \\
vs. Reflect on Ins. + Res.  & 68     & 97       & 53  & 1.069  \\
\hline
vs. Select by Randomness & 81     & 94       & 43 & 1.174  \\
vs. Select by Coherence & 75     & 96       & 47 &  1.128 \\
vs. Select by Perplexity & 64     & 99       & 55 & 1.041  \\ \hline
vs. Select by IFD only & 58     & 107       & 53  &  1.023 \\
vs. Select by r-IFD only & 74     & 96       & 48  &  1.119 \\
\hline
\end{tabular}
}
\caption{
The pair-wise comparison between our sRecycled WizardLM 7B with other models. The ``win'', ``Tie'' and ``Lose'' represent the number of wins or losses of sRecycled WizardLM 7B. The win rate is calculated as (Num(Win) $-$ Num(Lose)) $/$ Num(All) $+ 1$. 
}
\label{tbl:ablation}
\end{table}

\subsection{Ablation on Selection}

Moreover, to further verify the effectiveness of our selection mechanism, experiments with different selection methods are conducted shown in Table \ref{tbl:ablation}. 

``Select by Randomness'' represents the student model randomly choosing whether to accept improved data. Not only does this model underperform our final model largely, but it also underperforms both ``Reflect on Res.'' and ``Reflect on Ins. + Res.''. This baseline result indicates that without a proper selection method, the blind mixture of data might harm the model's performance. 

``Select by Coherence'' represents the data selected based on the coherence between instruction and response, which is calculated by cosine similarity of the Sentence-BERT \cite{reimers-gurevych-2019-sentence} embeddings. In this setting, the data pairs, whose instruction and response are more related, are more likely to be selected. The performance of this model is slightly better than the random selection model, and still worse than both ``Reflect on Res.'' and ``Reflect on Ins. + Res.'', indicating the ineffectiveness of this selection method. 

``Select by Perplexity'' represents the student model choosing whether to accept the improved data by whether the perplexity is improved, which is the closest to ours. The performance of this model surpasses both ``Reflect on Res.'' and ``Reflect on Ins. + Res.'', showing that a selection process can definitely further improve the model's performance, verifying our motivation for adding the selection mechanism. However, this model still underperforms our model, indicating the efficacy of our selection strategy.   

``Select by IFD only'' and ``Select by r-IFD only'' represent situations where we only utilize IFD or r-IFD scores for student side selection. Utilizing only IFD results in a model that is close to our main model, indicating the usefulness of the IFD score. However, its performance is still lower, indicating the effect of the r-IFD.

\section{ Comparison with Related Work  }

Earlier works on instruction tuning focus on creating large, high-quality datasets curated by human experts~\cite{khashabi-etal-2020-unifiedqa, ye-etal-2021-crossfit, wei2022finetuned, wang-etal-2022-super, du-etal-2022-glm}, time-consuming and labor-intensive. Thus a number of works try to construct instruction-tuning datasets automatically. 
Self-Instruct \cite{wang-etal-2023-self-instruct} utilizes the in-context learning capability of GPT-3 to expand tasks to many diverse instruction-response pairs. 
WizardLM \cite{xu2023wizardlm} applies an evolution methodology to refine and diversify the original instruction data. 
LaMini-LM \cite{wu2024laminilm} introduces to generate Top-Fuided instructions based on Wiki data. 
\citet{peng2023instruction} utilize GPT4 to generate responses for existing datasets. 
UltraChat \cite{ding2023enhancing}, establishes various scopes and systematically generates a multitude of instructions within each designated area. 
Orca \cite{mitra2023orca} directly apply GPT4 to generate reasoning steps for given instructions. 
SelFee \cite{selfee2023} utilizes ChatGPT to enhance the response quality. 
Reflection-Tuning \cite{li2023reflectiontuning} improves both the instruction and response sequentially by
reflecting on specific criteria. 
DEITA \cite{liu2023makes} utilizes ChatGPT to diversify and then select the data. 
LIFT \cite{xu2023rethinking} also tries to utilize ChatGPT/GPT4 to expand and compress the data. 

All the above works are related to ours by involving a teacher model to improve the instruction data, however, all of them are \textbf{teacher-dominating}: Both the generation and selection are all decided by the teacher model and without involving the student. We are the first to introduce the \textbf{teacher-student collaboration pipeline} and it works fine.

\section{Conclusion}

Selective Reflection-Tuning, as proposed in this paper, marks a significant advancement in data improvement for instruction tuning of Large Language Models. By integrating an interactive pipeline between a teacher model and a student model, and utilizing the novel metrics of IFD and reversed-IFD, this approach has demonstrated a marked improvement in the quality and relevance of instruction-tuning datasets. 
The resulting enhancement in model performance across various benchmarks not only attests to the efficacy of our method but also suggests its potential applicability in broader machine learning contexts. 

\section*{Limitations}

The involvement of the student model makes it possible to build high-quality and student-compatible instruction-response data. However, the main limitation of this method is that the data samples selected by different student models are different, thus the statistics (IFD scores and r-IFD scores) need to be calculated again for different student models. We believe the use of model-specific data samples is more reasonable due to the distinct characteristics of different models, and utilizing the statistics-based method is much more efficient than other generation-based methods, the necessity of re-calculation for new models is still not efficient enough.

\section*{Acknowledgement}

This work was supported in part by Adobe Research.

\bibliography{custom}

\appendix

\clearpage
\section{Prompt for Evaluation}
\label{p_evaluation}

We provide the detailed prompt we used for the pair-wise comparison in Figure \ref{appendix_prompt}. 

\begin{figure}[h]
  \centering
  \parbox{0.48\textwidth}{
        \rule{0.48\textwidth}{1.5pt} 
        Prompt for Performance Evaluation \\
        \rule{0.48\textwidth}{0.8pt} 
        \textbf{System Prompt} \\
        You are a helpful and precise assistant for checking the quality of the answer. \\

        \textbf{User Prompt} \\
        \text{[Question]}\\
        \textit{Question}\\
        \text{[The Start of Assistant 2's Answer]}\\
        \textit{Answer 2}\\
        \text{[The End of Assistant 2's Answer]}\\
        \text{[The Start of Assistant 2's Answer]}\\
        \textit{Answer 2}\\
        \text{[The End of Assistant 2's Answer]}\\

        We would like to request your feedback on the performance of two AI assistants in response to the user question displayed above. \\
        Please rate the helpfulness, relevance, accuracy, level of details of their responses. Each assistant receives an overall score on a scale of 1 to 10, where a higher score indicates better overall performance. \\
        Please first output a single line containing only two values indicating the scores for Assistant 1 and 2, respectively. The two scores are separated by a space. In the subsequent line, please provide a comprehensive explanation of your evaluation, avoiding any potential bias and ensuring that the order in which the responses were presented does not affect your judgment.

        \rule{0.48\textwidth}{0.8pt} 

  }
\caption{
The prompt we used to request ChatGPT or GPT4 to evaluate the responses. 
} 
\label{appendix_prompt} 
\end{figure}

\clearpage
\section{Prompt for Reflection}
\label{p_reflection}

The prompts for the reflection are shown in Figure \ref{prompt_Q} and Figure \ref{prompt_A}.

\begin{figure*}[h]
  \centering
  \parbox{0.98\textwidth}{
        \rule{0.98\textwidth}{1.5pt} 
        Prompt for Reflecting Instruction \\
        \rule{0.98\textwidth}{0.8pt} 
        \textbf{System Prompt} \\
        You are a helpful, precise but picky assistant for checking the quality of a given instruction. \\

        \textbf{User Prompt} \\
        \text{[Instruction]}\\
        \textit{Instruction}\\
        \text{[The Start of Answer]}\\
        \textit{Answer}\\
        \text{[The End of Answer]}\\

        We would like you to answer several questions related to the quality of a given instruction.\\
        1. Why this instruction is not good? First analyze the instruction based on the Complexity of the Topic, Level of Detail Required, Knowledge Required, Ambiguity of the Instruction and Logical Reasoning or Problem-Solving Involved. Then analyze why this answer is not good for the given instruction based on the Helpfulness, Relevance, Accuracy and Level of Details. Finally, analyze why this bad instruction leads to a bad answer.\\
        2. Based on the reason you provided, generate a new and complete instruction that is complex and difficult to answer directly. Make sure the new instruction is relevant but independent to the original instruction, which can be answered without knowing the original instruction, put the new instruction in the format of [New Instruction] your instruction [End]\\
        3. Answer the newly generated instruction as detailed as possible, in the format of [New Answer] your answer [End]

        \rule{0.98\textwidth}{0.8pt} 
    }
\caption{
The prompt we used to modify the existing instruction.
} 
\label{prompt_Q} 
\end{figure*}

\begin{figure*}[h]
  \centering
  \parbox{0.98\textwidth}{
        \rule{0.98\textwidth}{1.5pt} 
        Prompt for Reflecting Response \\
        \rule{0.98\textwidth}{0.8pt} 
        \textbf{System Prompt} \\
        You are a helpful, precise but picky assistant for checking the quality of the answer to a given instruction. \\

        \textbf{User Prompt} \\
        \text{[Instruction]}\\
        \textit{Instruction}\\
        \text{[The Start of Answer]}\\
        \textit{Answer}\\
        \text{[The End of Answer]}\\

        We would like you to answer several questions related to the quality of the answer to the given instruction.\\
        1. Why this answer is not good for the given instruction? Analyze based on the Helpfulness, Relevance, Accuracy, and Level of Details.\\
        2. Based on the reason you provided, generate a better answer, new and complete, as detailed as possible, in the format of [Better Answer] your answer [End] 

        \rule{0.98\textwidth}{0.8pt} 
    }
\caption{
The prompt we used to modify the existing response.
} 
\label{prompt_A} 
\end{figure*}

\clearpage
\section{Evaluation Metric}
\label{evaluation}

\subsection{Pair-wise comparison}

Evaluation of the responses generated by LLMs is an open problem that plenty of researchers are still working on, due to the lack of real ground truth for the open-domain questions, most of the previous methods can not be directly implemented for judging the instruction-following ability of LLMs. However, using LLM as a judge, e.g., GPT4, for evaluation is recently a widely accepted and common practice \cite{touvron2023llama2, vicuna2023, dettmers2023qlora, liu2023geval, chiang-lee-2023-large}. Previous studies \cite{zheng2023judging, alpaca_eval} have shown that GPT4's evaluations are consistent with human evaluations. We utilized the testing instruction set from WizardLM \cite{xu2023wizardlm} which contains $218$ diverse human-curated instructions, which are categorized into specific sub-categories.

Specifically, we directly follow the evaluation method from \citet{chen2023alpagasus, li2023quantity}, which contains rating each model-generated response on a scale spanning from \(1\) to \(10\), with scores encapsulating several aspects such as accuracy and relevance. To further mitigate the positional bias elaborated upon in \cite{ko-etal-2020-look, wang2023large}, model-generated outputs are presented to the LLM judge in two distinct sequences and subsequently scored. Hence, a model's dominance is ratified under the following conditions: 
\textbf{Wins:} Exhibits superiority in both sequences or prevails in one while maintaining parity in the alternate sequence.
\textbf{Tie:} Demonstrates parity across both sequences or prevails in one while faltering in the alternate.
\textbf{Loses:} Underperforms in both sequences or maintains parity in one while being eclipsed in the alternate.

\subsection{Alapca Eval Leaderboard}

AlpacaEval Leaderboard offers an LLM-centric automatic assessment utilizing the AlpacaFarm \cite{dubois2023alpacafarm} evaluation dataset. It is an automated evaluation mechanism for LLMs that offers efficiency, cost-effectiveness, and reliability. Operating on the AlpacaFarm evaluation dataset, it gauges models' proficiency in adhering to generic user instructions. The generated outputs are juxtaposed against benchmark responses from Davinci003. Empirical evidence suggests that AlpacaEval's alignment with ground truth annotations sourced from human experts is notably high.

\subsection{Open LLM Leaderboard}

The Huggingface Open LLM Leaderboard employs the evaluation methodology from \cite{eval-harness}, providing a cohesive framework for assessing generative language model capabilities across a spectrum of evaluation tasks. It focuses on $4$ pivotal benchmarks: ARC \cite{clark2018think}, HellaSwag \cite{zellers-etal-2019-hellaswag}, MMLU \cite{hendrycks2021measuring}, and TruthfulQA \cite{lin-etal-2022-truthfulqa}.

\subsection{MT-Bench}

We also provide the performances of our sRecycled Models on MT-bench, as shown in Table \ref{tbl:mt_bench}. Since our training focused on 1-turn instructions and did not include any multi-turn data, the 1-turn score on the MT bench is promising and comparable to LLaMA2-13B-chat, while the 2-turn score is not that satisfactory. 
However, the Vicuna dataset \citet{vicuna2023} can introduce multi-turn dialog data to the model training. Hence, we tried training with our data based on the existing Vicuna 7B v1.5 model, whose result is reported in the last row as “sRecycled Wiz + Vicuna 7B”. Compared with the original Vicuna model, the 1-turn, 2-turn, and overall scores are improved dramatically and the overall score is similar to the performance of Vicuna-13B.

\begin{table}[!tbh]
\centering
\scalebox{0.9}{
\begin{tabular}{lccc}
\hline
& 1-turn & 2-turn & Overall \\\hline
sRecycled Alpaca 7B & 6.653 & 2.888  & 4.891 \\
sRecycled Wiz 7B & 6.538 & 4.588 & 5.563 \\ \hline
Vicuna 7B v1.5 & 6.569 & 5.588  & 6.078 \\
sRecycled Wiz + Vicuna 7B & 7.063 & 5.975 & 6.519 \\

\hline
\end{tabular}
}
\caption{
The MT-Bench results of our models, including 1-turn, 2-turn, and Overall Scores. 
}
\label{tbl:mt_bench}
\end{table}

\subsection{Human Study}

To further validate the superiority of our method, we conducted a further human study to further evaluate the effectiveness of our method. In the test set, there are $27$ sub-categories that have $4$ or more testing instructions, thus we randomly sampled $4$ instructions from each sub-category to form a set containing $108$ instructions. Then $3$ human participants are given the task of comparing the responses generated by the comparing models with the criteria same as the previous pair-wise evaluation. For each comparison, 3 options are given (Win, Tie, and Loss) and the final results are determined by the majority voting of the participants. 

\clearpage
\section{Implementation Details}
\label{implement}

For the Llama2 pre-trained model \cite{touvron2023llama2}, we utilize the prompt and code base from Vicuna \cite{vicuna2023} and flash attention \cite{dao2022flashattention} while the overall training arguments are aligned with protocols from Alpaca and WizardLM datasets. The Adam optimizer \cite{kingma2017adam}, with a $2\times10^{-5}$ learning rate for the 7b model and a $1\times10^{-5}$ learning rate for the 13b model, and a batch size of $128$, steer the training across three epochs with a max length of $2048$. The warmup rate is set to $0.03$. 

\clearpage
\section{Statistic Analysis}
\label{statistic}

\subsection{Basic Data Statistics}

In this section, we delve into a quantitative analysis of the instruction-response data, pre- and post-application of our methodology, as delineated in Table \ref{tbl:comparison}. We first compare both ``Recycled Data'' and ``sRecycled Data'' to the original data. 

Observationally, there's an increase in the average token length of instructions within the Alpaca dataset, whereas a decrement manifests for the WizardLM dataset, epitomizing the method's adept adaptability. The succinctness and elementary nature of the Alpaca dataset's instructions warrant an enhancement in intricacy through our method, thereby elongating their length. Conversely, the pre-existing complexity and intricacy in WizardLM's instructions render our algorithm inclined towards succinctness. 
Pertaining to the response section, there's a marked propensity of our approach to engender detail-rich textual content, leading to relatively long responses. 

Moreover, leveraging Sentence-BERT \cite{reimers-gurevych-2019-sentence}, we quantify the coherence metric between instructions and their affiliated responses. It's discernible that our technique invariably fabricates samples with better coherence, signifying a superior alignment between modulated instructions and consequent responses. 
Additionally, to elucidate the metamorphosis in instructional difficulty, we employ the IFD score, executed on the pre-trained llama2-7b language model to check the the difficulties of instructions. The increase in IFD scores represents the increase in the overall difficulty of instructions. Moreover, r-IFD is also calculated, and the decrease in r-IFD scores represents the instruction response pair is more related.

\begin{table*}[h]
\centering
\scalebox{0.85}{
\begin{tabular}{l|cccccccc}
\hline
& \multicolumn{7}{c}{Comparison of Different Models}  \\
& Ins. len & Res. len& Ins. ppl & Res. ppl 1 & Res. ppl 2 & Coherent & IFD  & r-IFD  \\
\hline
Original Alpaca Data& 20.7 & 65.5 & 34.3 & 82.6 & 49.2 & 0.53 & 0.71 & 0.48\\
Recycled Alpaca Data& 37.9 & 377.2 & 13.6 & 4.5 & 2.9 & 0.67 & 0.84 & 0.50\\
\textbf{sRecycled Alpaca Data}& 31.4 & 345.9 & 19.8 & 4.2 & 2.8 & 0.65 & 0.84 & 0.36\\
\hline
Original WizardLM Data& 123.0 & 348.5 & 12.3 & 17.0 & 7.5 & 0.65 & 0.71 & 0.52\\
Recycled WizardLM Data& 66.9 & 518.7 & 10.0 & 3.2 & 2.5 & 0.73 & 0.83 & 0.41\\
\textbf{sRecycled WizardLM Data}& 70.7 & 519.6 & 12.0 & 3.1 & 2.4 & 0.72 & 0.83 & 0.39\\
\hline
\end{tabular}}
\caption{
The comparison of some basic statistics. ``Ins. len'' and ``Res. len'' represent the average token length of the instructions and responses. ``Ins. ppl'' represents the average perplexity of instructions. ``Res. ppl 1'' and ``Res. ppl 2'' represent response perplexities without or with the context of corresponding instructions. All the perplexity is calculated upon our initial pre-trained model llama2-7b. ``Coherent'' represents the coherent score calculated by SentenceBert. ``IFD'' represents the instruction-following difficulty score proposed by Cherry LLM \cite{li2023quantity} and ``r-IFD'' represents the reversed instruction-following difficulty score proposed by us. 
}
\label{tbl:comparison}
\end{table*}

\subsection{Data Component Distribution}

\begin{figure*}[!htbh]
\centering 
\includegraphics[width=1\textwidth]{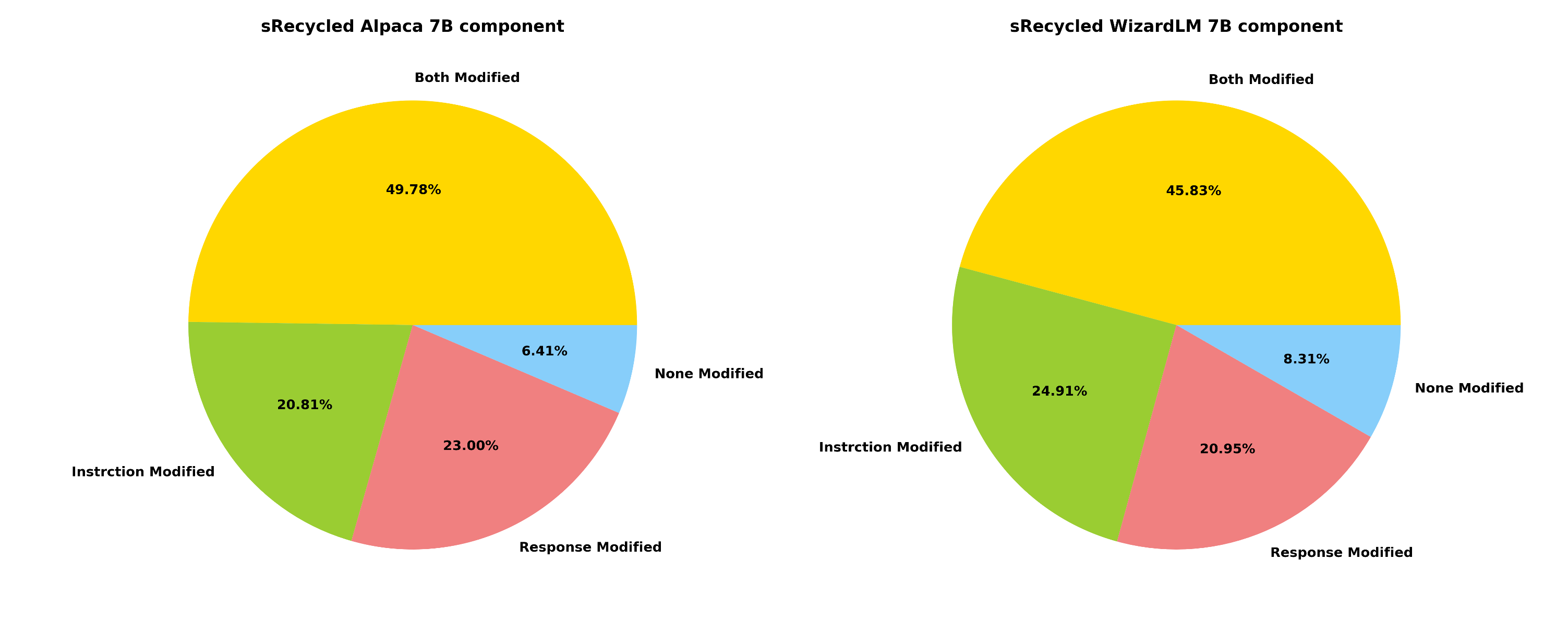} 
\caption{
The component distribution of the sRecycled Alpaca 7B and  sRecycled Wizardlm 7B data. 
} 
\label{pie} 
\end{figure*}

In our selective reflection-tuning, there are four different outcomes for each original data sample: both instruction and response are modified, only instruction is modified, only response is modified, and none of instruction and response are modified. Thus to provide a better view of the data conponents, we provide the pie chart for our sRecycled Alpaca 7B and sRecycled Wizardlm 7B data as shown in Figure \ref{pie}. 

\clearpage
\section{Detailed Few Data Scenario}
\label{few}

The detailed performances in the few data scenarios are shown in TABLE \ref{tbl:bench_few} and comparisons with the randomly selected method are shown in TABLE \ref{tbl:ablation_few}. 

\begin{table*}[!htbh]
\centering
\scalebox{0.85}{
\begin{tabular}{l|ccccc|c}
\hline
& \multicolumn{5}{c|}{\textbf{Huggingface Open LLM Leaderboard}} & \textbf{AlpacaEval} \\
& \textbf{Average} & \textbf{ARC} & \textbf{HellaSwag} & \textbf{MMLU} & \textbf{TruthfulQA} & \textbf{AlpacaEval} \\
\hline
{sRecycled WizardLM 7B (1\%) (463)} & 57.31 & 54.86 & 78.40 & 46.17 & 49.79 & 67.79  \\
{sRecycled WizardLM 7B (2\%) (926)} & 57.80 & 54.69 & 78.80 & 47.00 & 50.70 & 74.29  \\
{sRecycled WizardLM 7B (3\%) (1,390)} & 57.34 & 55.12 & 78.80 & 42.68 & 49.16 & 74.50  \\
{sRecycled WizardLM 7B (5\%) (2,316)} & 57.91 & 54.86 & 79.83 & 46.69 & 50.23 & 77.78  \\
{sRecycled WizardLM 7B (10\%) (4,632)} & 57.71 & 55.46 & 79.56 & 46.83 & 48.98 & 78.43  \\
{sRecycled WizardLM 7B (30\%) (13,897)} & 56.89 & 54.61 & 79.25 & 44.67 & 49.05 & 82.48  \\
{sRecycled WizardLM 7B (50\%) (23,163)} & 56.98 & 55.11 & 78.87 & 45.31 & 48.63 & 81.47  \\
{sRecycled WizardLM 7B (70\%) (32,428)} & 56.63 & 54.95 & 78.55 & 46.31 & 46.71 & 81.47  \\
\hline
{sRecycled WizardLM 7B (100\%) (46,325)} & 56.79 & 54.78 & 77.86 & 45.63 & 48.91 & 83.21 \\
\hline
\end{tabular}
}
\caption{
The comparison of performance on Huggingface Open LLM Leaderboard and AlpacaEval Leaderboard by using different amounts of selective recycled WizardLM data. In the first parentheses are the percentage of data used for tuning and in the second parentheses are the specific amount of number used.  
}
\label{tbl:bench_few}
\end{table*}

\begin{table*}[!htbh]
\centering
\scalebox{0.85}{
\begin{tabular}{l|ccccc|c}
\hline
& \multicolumn{5}{c|}{\textbf{Huggingface Open LLM Leaderboard}} & \textbf{AlpacaEval} \\
& \textbf{Average} & \textbf{ARC} & \textbf{HellaSwag} & \textbf{MMLU} & \textbf{TruthfulQA} & \textbf{AlpacaEval} \\
\hline
{sRecycled Wiz 7B (2\%) (926) (IFD)} & 57.80 & 54.69 & 78.80 & 47.00 & 50.70 & 74.29  \\
{sRecycled Wiz 7B (2\%) (926) (Random)} & 56.13 & 54.77 & 78.98 & 43.15 & 47.64 & 72.13  \\ \hline
{sRecycled Wiz 7B (5\%) (2,316) (IFD)} & 57.91 & 54.86 & 79.83 & 46.69 & 50.23 & 77.78  \\
{sRecycled Wiz 7B (5\%) (2,316) (Random)} & 57.07 & 54.10 & 78.97 & 46.54 & 48.67 & 76.40  \\ \hline
{sRecycled Wiz 7B (10\%) (4,632) (IFD)} & 57.71 & 55.46 & 79.56 & 46.83 & 48.98 & 78.43  \\
{sRecycled Wiz 7B (10\%) (4,632) (Random)} & 57.06 & 54.86 & 78.09 & 46.82 & 48.46 & 77.11  \\ \hline
{sRecycled Wiz 7B (30\%) (13,897) (IFD)} & 56.89 & 54.61 & 79.25 & 44.67 & 49.05 & 82.48  \\
{sRecycled Wiz 7B (30\%) (13,897) (Random)} & 56.80 & 54.95 & 78.07 & 47.39 & 46.81 & 79.73  \\

\hline
\end{tabular}
}
\caption{
The comparison of performance on Huggingface Open LLM Leaderboard and AlpacaEval Leaderboard by using different strategies in the few data scenarios.  
}
\label{tbl:ablation_few}
\end{table*}

\clearpage
\section{Ablation on Larger Evaluate Set}
\label{ablation}

The evaluation set used on the main page in Table \ref{tbl:ablation} is the WizardLM test set, which contains 218 human-written instructions, and is currently one of the most widely used test sets. Another widely used test set is the Vicuna test set, which is used in MT-Bench, but it contains only 80 instructions and the results are presented in Appendix \ref{evaluation}. Thus the test set we used for ablation is almost three times the Vicuna set. Moreover, in our evaluation, every comparison will be processed twice to eliminate the potential position bias. Thus we don't think it would be regarded as a really small test set. 

However, to further validate the effectiveness of our method, we further combine the Vicuna \cite{vicuna2023} test set (80), Koala \cite{vu2023koala} test set (180), WizardLM \cite{xu2023wizardlm} test set (218), Self-instruct \cite{wang-etal-2023-self-instruct} test set (252), and LIMA \cite{zhou2023lima} test set (300) into a huge evaluation set of 1030 instructions for the ablation study as shown in Table \ref{tbl:ablation_large}. The results on this huge test set share similar trends compared with using the WizardLM test set alone, indicating the effectiveness of our method.

\begin{table}[!tbh]
\centering
\scalebox{0.8}{
\begin{tabular}{lcccc}
\hline
& Win & Tie & Lose & Win Rate\\\hline
vs. Original WizardLM & 738 & 126 & 166 & 1.556 \\
\hline
vs. Reflect on Ins.   & 688 & 196 & 146 & 1.548 \\
vs. Reflect on Res.   & 360 & 388 & 282 & 1.078 \\
vs. Reflect on Ins. + Res.  & 329 & 434 & 267 & 1.044 \\
\hline
vs. Select by Randomness & 401 & 416 & 213 & 1.149 \\
vs. Select by Coherence & 360 & 444 & 226 & 1.108 \\
vs. Select by Perplexity & 307 & 438 & 285 & 1.050  \\
\hline
\end{tabular}
}
\caption{
The pair-wise comparison between our sRecycled WizardLM 7B with other models. The ``win'', ``Tie'' and ``Lose'' represent the number of wins or losses of sRecycled WizardLM 7B. The win rate is calculated as (Num(Win) $-$ Num(Lose)) $/$ Num(All) $+ 1$. 
}
\label{tbl:ablation_large}
\end{table}

\clearpage
\section{Examples for r-IFD Illustrtaion}
\label{rifd}

\rule{0.44\textwidth}{0.8pt}

\noindent
\textbf{Example 1:} \\
(r-IFD=0.148, Low, Prefered)

\noindent
\textbf{Instruction}: \\
Reverse a linked list.

\noindent
\textbf{Response}: 
\begin{lstlisting}
def reverse(head): 
    prev = None
    cur = head 

    while cur:
        next = cur.next
        cur.next = prev 
        prev = cur 
        cur = next
    head = prev
    
    return head
\end{lstlisting}
\rule{0.44\textwidth}{0.8pt}

\noindent
\textbf{Example 2:} \\
(r-IFD=0.921, High, Not Prefered)

\noindent
\textbf{Instruction}: \\
 Identify the type of sentence "I drove to the store yesterday".

\noindent
\textbf{Response}: \\
Declarative sentence.

\rule{0.44\textwidth}{0.8pt}

In the first example, after reading through the given code, LLM can easily understand the task and guess what this code is for, indicating sufficient information in the response and its good match to the instruction. However, in the second example, the response is not able to provide enough information to derive the instructions and is vague in various aspects. It indicates that the response might not be feasible to be reasoned by the model and thus needs to be improved.

\end{document}

%% file: introduction.tex
\section{Introduction}

The quality of instruction tuning \cite{wei2022finetuned, chen2023alpagasus, mishra2021cross, Chung2022ScalingIL, zhang2023instruction, liu2023mmc} data is paramount to the LLM being fine-tuned, i.e., the student model. There is a growing trend and demand for the community to automatically improve the quality of instruction tuning data. Previous works either curate datasets by human experts \cite{DatabricksBlog2023DollyV2, Longpre2023TheFC, zhou2023lima} or distill the responses of well-trained LLMs \cite{alpaca, peng2023instruction, vicuna2023, vu2023koala, xu2023wizardlm, li2023reflectiontuning, Li2024CanLS, xu2024survey}. The self-improvement \cite{bai2022constitutional, self-improve, pan2023automatically} ability of LLMs has also been explored to improve the instruction or response of a training sample.

However, these existing methods of data enhancement \cite{self-improve, selfee2023, li2023reflectiontuning, mitra2023orca} do not take a critical criterion into account: \textbf{Is the teacher-refined data compatible to the needs of the student model?} These approaches typically do not account for the inherent randomness and potential degradation associated with the generative models' output, leading to an oversight in how the student model responds to these ``improved'' data samples. Thus a mechanism for the student model to selectively integrate these enhancements has been notably absent. 
To bridge this gap, our work introduces an \textbf{teacher-student collaboration pipeline} wherein a teacher generative model engages in a reflection process to enhance both the instruction and response of a data sample. The student model then evaluates whether to incorporate these improvements based on its unique statistical attributes. This pipeline is versatile and can be adapted to various contexts where data enhancement is needed.

Then, another pivotal question arises: \textbf{How does the student model decide which enhanced data are needed and critical to its training?} This question underpins the challenge of autonomously evaluating the quality of instructions and responses. Common practices involve utilizing sophisticated models like GPT-4 for assessment purposes \cite{zheng2023judging, alpaca_eval, liu2023geval, chiang-lee-2023-large} or employing a secondary judge model equipped with evaluative capabilities \cite{wang2023shepherd, li2023generative}. These methods, however, present limitations: they fail to address the discrepancies between the evaluating model and the actual student model undergoing training. Particularly in the latter approach, even though the judge model and the student model might share the same structural framework, their weight distributions diverge once endowed with the evaluative functions. Consequently, the preferences of the judge model may not align with the real student model's requirements. To circumvent these issues, we adopt a statistical method, utilizing the Instruction-Following Difficulty (IFD) score proposed by \citet{li2023quantity, Li2024SuperfilteringWD}. This score is derived directly from the raw student model, thereby mitigating potential domain shifts and ensuring that the evaluation is better aligned with the student model’s learning context. 

In our approach, the IFD score serves as a crucial metric that measures how much help the instruction can provide to the likelihood of the response if added as an extra condition, representing the \textbf{Difficulty} of the sample. 
However, though effective, the IFD score mainly assesses the instructions. Motivated by Humpback \cite{li2023self} which requires LLMs to generate potential instruction based on responses, we further introduce a reversed version of IFD named reversed-IFD (r-IFD). This metric evaluates how much the response contributes to predicting the corresponding instruction. A lower r-IFD score suggests the student can easily deduce the corresponding instruction given the response, indicating this sample is feasible for the student to learn, representing the \textbf{Feasibility} of the sample. 
This dual approach, employing both \textbf{IFD scores for Difficulty} and \textbf{r-IFD scores for Feasibility}, enables a comprehensive and nuanced assessment of the instruction-tuning process, ensuring the refined data aligns well with the student model’s capabilities and objectives.

We name our overall method Selective Reflection-Tuning, which contains the selective instruction reflection phase and the selective response reflection phase. In the first phase, a teacher model is utilized to reflect on the instruction of the given sample based on some criteria and generate a new sample. Then the student model makes the decision of whether to accept the improvement based on difficulty (IFD). In the second phase, the teacher model reflects and generates a sample with a new response and the student model decides whether to accept based on feasibility (r-IFD). With our interactive pipeline, we obtain a dataset with supreme quality, with only instruction tuning on a relatively small amount of data, our model outperforms most existing open-source models with even larger model sizes. \looseness-1 
Our contributions include:
\vspace{-2mm}
\begin{itemize}
    \item We propose a teacher-student collaboration pipeline where the teacher model and student model cooperate to build a more coherent and model-compatible instruction tuning dataset, which can be further adapted into other self-improvement scenarios. \vspace{-2mm}
    \item We present a nuanced evaluation schema reversed-IFD, quantifying the relevance of instruction-response pairs, and representing the feasibility of the sample for the student. \vspace{-2mm}
    \item With only instruction tuning on a few thousand of automatically generated data, our models achieve top-tier performances, indicating the supreme quality of our data. \vspace{-2mm}
\end{itemize}

%% file: method.tex
\begin{figure*}[!htbh]
\centering 
\includegraphics[width=0.99\textwidth]{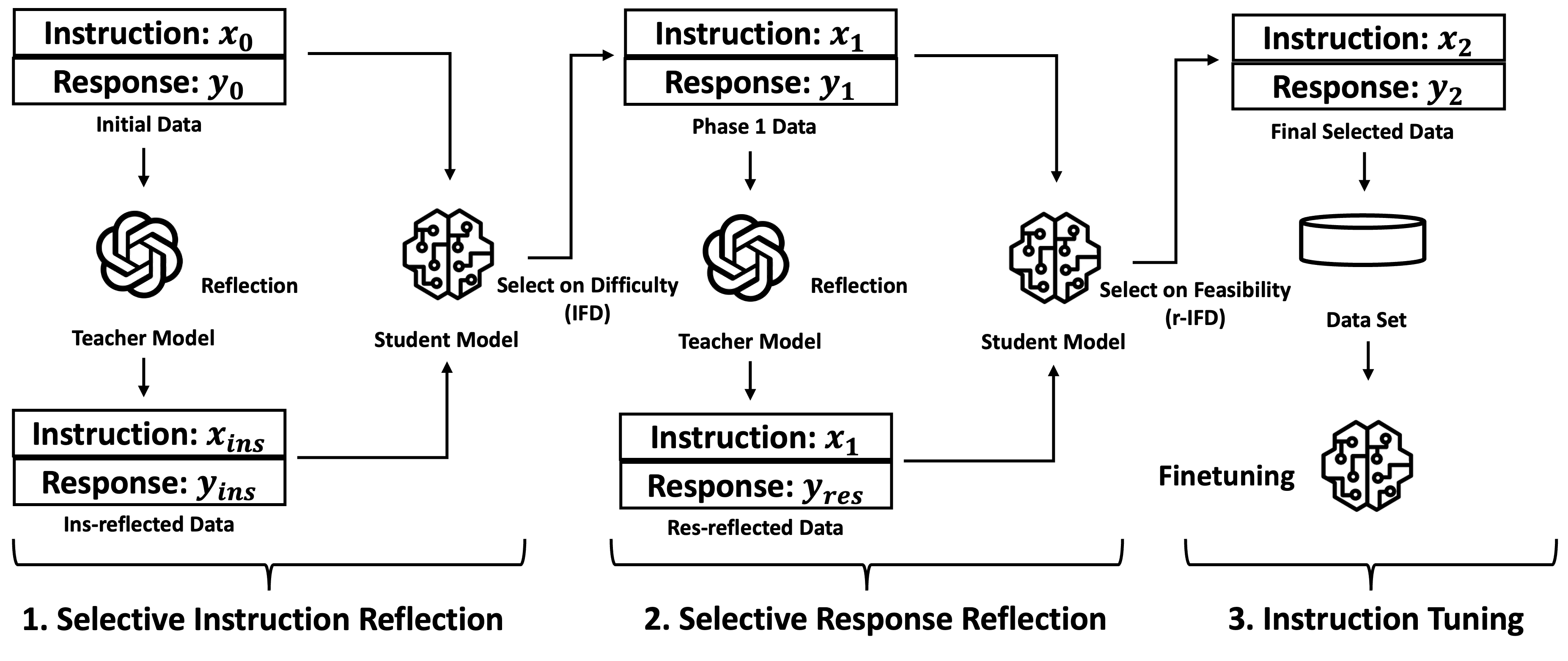} 
\caption{
The overall pipeline of our method. The first Selective Instruction Reflection phase aims to obtain a better instruction for a data sample and the second Selective Response Reflection phase aims to obtain a better response for the sample. The reflection process is conducted by the well-trained teacher model and the selection process is conducted by the student model. 
} 
\label{method} 
\vspace{-2mm}
\end{figure*}

\section{Preliminaries}

Let $f_\theta$ denote the pre-trained student model, e.g., LLaMA, with parameters $\theta$ and $g$ the teacher model, e.g., ChatGPT. Let lowercase letters $x, y, z, c,..$ denote the text segments, which could be phrases or sentences, and each token in $x$ is denoted as $x[i]$. We use uppercase letters $D, ..$ to denote the collection of language sequences or datasets, and $D_0$ represents the initial base dataset. Since both $f_\theta$ and $g$ are in auto-regressive manners, a sequence $x = (x[1],...,x[n])$ can be further denoted as:
\begin{equation}
    f_\theta(x) = \prod_{i=1}^{n} f(x[i] | x[1,...,i-1])
\end{equation}

In the instruction tuning setting, there will be a mapping function that turns the original raw instruction $x$ into the desirable format and requests models for a response $y$. For simplicity, we directly notate this process as $y \sim f(y|x)$. And the loss function for instruction-tuning can be denoted as:
\begin{equation}
    L_\theta(y|x) = -\frac{1}{n}\sum_{i=1}^{n} \log f_\theta(y|x)
\end{equation}
where $n$ is the length of response $y$.

Motivated by Cherry LLM \cite{li2023quantity} which proposes the IFD score to measure the difficulty of instruction in the given instruction-response pairs. We utilize the perplexity of the IFD score \cite{Li2024SuperfilteringWD}, which is formulated as:
\begin{align}
     \text{IFD}_\theta(y|x) & = \frac{\text{ppl}(y|x)}{\text{ppl}(y)} = \exp (L_\theta(y|x) - L_\theta(y))
\end{align}
where $\text{ppl}(y|x)$ represents the perplexity of model $f_\theta$ to fit the response $y$ given the instruction $x$ as the context, and $\text{ppl}(y)$ represents the perplexity of model $f_\theta$ to directly fit the response $y$ without any context given. This value represents how the given instruction $x$ affects the generation of corresponding response $y$ for given model $f_\theta$, which has been shown as an effective metric for evaluating the given instruction-following data pairs \cite{Li2024SuperfilteringWD}. A higher IFD score indicates that the instruction is more challenging for the student model to generate the response, suggesting the instruction's difficulty for the student model. 

\section{Methodology}

As shown in Figure \ref{method}, there are two main phases in our method, Selective Instruction Reflection and Selective Response Reflection phase. In each phase, the teacher model generates the updated version of instructions or responses based on some given specific criteria $\{c_{ins, 1}, ..., c_{ins, k}\}$ \footnote{Prompt for reflection can be found in Appendix \ref{p_reflection}}, then the student model judges if the updates are beneficial to it based on difficulty (IFD) or feasibility (reverse-IFD). Finally, these selectively improved samples can be used for the final instruction tuning.

\subsection{Selective Reflection on Instruction}

\noindent
\textbf{Reflection on Instruction}

Given the instruction-response pair $(x_0, y_0)$ from the original dataset $D_0$ with some specific criteria $\{c_{ins, 1}, ..., c_{ins, k}\}$, the teacher model $g$ is required to reflect on this sample and generate a better instruction-response pair $(x_{ins}, y_{ins})$ according to its reflection.
With the criteria given, the teacher model $g$ is able to generate critical responses: 
\begin{equation}
     [ z_{ins, 1}, ... ] \sim g(z, ...|x_0, y_0, c_{ins, 1}, ...)
\end{equation}
where both original instruction and response are wrapped into the prompt rather than original instruction alone. These critical responses further serve as the guidance (chain of thought) \cite{wei2023chainofthought, yao2023tree} for the generation of the new instruction and response pair: 
\begin{equation}
     [ x_{ins}, y_{ins} ] 
     \sim g( x, y|x_0, y_0, c_{ins, 1}, ..., z_{ins, 1}, ...)
\end{equation}
where the above process is sampled as a continuous language sequence, and the critical responses would not be decomposed from the whole output.

\noindent
\textbf{Selection on Instruction}

Though the given sample pair is updated by the teacher model, it remains uncertain whether this updated version is truly better for the student model. While most existing work evaluates the quality of a data sample by directly prompting existing generative models, they inevitably suffer from the misalignment problem. Thus we utilize the IFD score \cite{li2023quantity} calculated based on the specific base student model, which measures how the instruction benefits the generation of corresponding responses for the model, representing the difficulty of the sample. 

After obtaining the updated instruction-response pair, the base model $f_\theta$ is required to compare the IFD score of the original pair $(x_{0}, y_{0})$ and updated pair $(x_{ins}, y_{ins})$ and the sample with higher IFD scores will be chosen:  
\begin{equation}
(x_1, y_1) = \underset{(x, y)}{\text{argmax}}(\text{IFD}_\theta(y|x))
\end{equation}
where $(x, y) \in \{(x_{0}, y_{0}), (x_{ins}, y_{ins})\}$. 
Then the chosen data pair $(x_1, y_1)$ with a higher IFD score will be sent to the next phase. 

\subsection{Selective Reflection on Response}

\noindent
\textbf{Reflection on Response}

After the first phase, although the instruction $x_{1}$ is guaranteed to be difficult for the student model, the corresponding response $y_{1}$ is still sub-optimal. Thus another reflection on the response process is further proposed. Similar to the above procedure, a new set of criteria for reflection on response is defined as $\{c_{res, 1}, ..., c_{res, m}\}$. The overall process can be noted as: 
\begin{equation}
    y_{res} \sim g(y|x_{1}, y_{1}, c_{res, 1}, ..., c_{res, m}, z_{res, 1}, ..., z_{res, m})
\end{equation}
where $z_{res, i}$ represents the critical response of $i$th response criteria $c_{res, i}$. In the process, the instruction and response pair $(x_{1}, y_{res}))$ is fully improved. 

\noindent
\textbf{Selection on Response}

Our pipeline aims to improve both the instruction and response in an instruction-tuning sample. IFD score measures the difficulty of the sample. We take a step further by adding another dimension which we call reversed IFD (r-IFD) representing the feasibility for the student to generate the instruction given the response. 
A lower r-IFD score suggests the student can easily deduce the corresponding instruction given the response, indicating this sample is feasible for the student to learn, which measures the model-specific matching degree of the existing data pair. \footnote{ Two examples with low or high r-IFD scores can be found in Appendix \ref{rifd} for better illustration. }

The high-level idea of r-IFD is in line with the success of Humpback \cite{li2023self}, which utilizes LLM to predict the corresponding instruction from given texts (responses), and hypothesizes that “we can predict instructions for these candidate gold answers that can be used as high-quality example pairs”. In our paper, we further hypothesize that a response is more informative for training if it is feasible for the LLM to predict the corresponding instruction from the response. This hypothesis is naturally proved by the Humpback, which generates instructions that can be handled by LLMs, while those difficult ones are naturally discarded. 

Under this circumstance, the reversed IFD score should be small since the smaller value represents that it is easier for the model to generate the corresponding instruction given the response. 
Specifically, the r-IFD score is calculated as: 
\begin{align} 
     \text{r-IFD}_\theta(x|y) & = \frac{\text{ppl}(x|y')}{\text{ppl}(x)} = \exp (L_\theta(x|y') - L_\theta(x))
\end{align}
where $y'$ represents the text segment generated by mapping the original $y$ into a query to guess the corresponding potential instructions.  

For the given original sample pair $(x_1, y_1)$ from the first phase and reflected sample pair $(x_1, y_{res})$, the selection process can be formulated as: 
\begin{equation}
(x_2, y_2) = \underset{(x, y)}{\text{argmin}}(\text{r-IFD}_\theta(x|y))
\end{equation}
where $(x, y) \in \{(x_{1}, y_{1}), (x_{1}, y_{res})\}$. 

After the above phases, there will be a corresponding data pair $(x_2, y_2)$ for each original $(x_0,y_0)$, which is represented as our selective reflected data. Then we discard all the samples which is not response-reflected for the consistency of response distribution. 
We name the whole above process as a selective recycling process, which greatly improves the quality of the previous dataset \footnote{Some statistic analysis can be found in Appendix \ref{statistic}}. The student model $f_\theta$ will be trained on the newly generated data and the new models are notated as ``sRecycled Models'', eg. sRecycled Alpaca.

%% file: acl_latex.bbl
\begin{thebibliography}{63}
\expandafter\ifx\csname natexlab\endcsname\relax\def\natexlab#1{#1}\fi

\bibitem[{Bai et~al.(2022{\natexlab{a}})Bai, Jones, Ndousse, Askell, Chen, DasSarma, Drain, Fort, Ganguli, Henighan, Joseph, Kadavath, Kernion, Conerly, El-Showk, Elhage, Hatfield-Dodds, Hernandez, Hume, Johnston, Kravec, Lovitt, Nanda, Olsson, Amodei, Brown, Clark, McCandlish, Olah, Mann, and Kaplan}]{bai2022training}
Yuntao Bai, Andy Jones, Kamal Ndousse, Amanda Askell, Anna Chen, Nova DasSarma, Dawn Drain, Stanislav Fort, Deep Ganguli, Tom Henighan, Nicholas Joseph, Saurav Kadavath, Jackson Kernion, Tom Conerly, Sheer El-Showk, Nelson Elhage, Zac Hatfield-Dodds, Danny Hernandez, Tristan Hume, Scott Johnston, Shauna Kravec, Liane Lovitt, Neel Nanda, Catherine Olsson, Dario Amodei, Tom Brown, Jack Clark, Sam McCandlish, Chris Olah, Ben Mann, and Jared Kaplan. 2022{\natexlab{a}}.
\newblock \href {http://arxiv.org/abs/2204.05862} {Training a helpful and harmless assistant with reinforcement learning from human feedback}.

\bibitem[{Bai et~al.(2022{\natexlab{b}})Bai, Kadavath, Kundu, Askell, Kernion, Jones, Chen, Goldie, Mirhoseini, McKinnon, Chen, Olsson, Olah, Hernandez, Drain, Ganguli, Li, Tran-Johnson, Perez, Kerr, Mueller, Ladish, Landau, Ndousse, Lukosuite, Lovitt, Sellitto, Elhage, Schiefer, Mercado, DasSarma, Lasenby, Larson, Ringer, Johnston, Kravec, Showk, Fort, Lanham, Telleen-Lawton, Conerly, Henighan, Hume, Bowman, Hatfield-Dodds, Mann, Amodei, Joseph, McCandlish, Brown, and Kaplan}]{bai2022constitutional}
Yuntao Bai, Saurav Kadavath, Sandipan Kundu, Amanda Askell, Jackson Kernion, Andy Jones, Anna Chen, Anna Goldie, Azalia Mirhoseini, Cameron McKinnon, Carol Chen, Catherine Olsson, Christopher Olah, Danny Hernandez, Dawn Drain, Deep Ganguli, Dustin Li, Eli Tran-Johnson, Ethan Perez, Jamie Kerr, Jared Mueller, Jeffrey Ladish, Joshua Landau, Kamal Ndousse, Kamile Lukosuite, Liane Lovitt, Michael Sellitto, Nelson Elhage, Nicholas Schiefer, Noemi Mercado, Nova DasSarma, Robert Lasenby, Robin Larson, Sam Ringer, Scott Johnston, Shauna Kravec, Sheer~El Showk, Stanislav Fort, Tamera Lanham, Timothy Telleen-Lawton, Tom Conerly, Tom Henighan, Tristan Hume, Samuel~R. Bowman, Zac Hatfield-Dodds, Ben Mann, Dario Amodei, Nicholas Joseph, Sam McCandlish, Tom Brown, and Jared Kaplan. 2022{\natexlab{b}}.
\newblock \href {http://arxiv.org/abs/2212.08073} {Constitutional ai: Harmlessness from ai feedback}.

\bibitem[{Chen et~al.(2023{\natexlab{a}})Chen, Li, Yan, Wang, Gunaratna, Yadav, Tang, Srinivasan, Zhou, Huang, and Jin}]{chen2023alpagasus}
Lichang Chen, Shiyang Li, Jun Yan, Hai Wang, Kalpa Gunaratna, Vikas Yadav, Zheng Tang, Vijay Srinivasan, Tianyi Zhou, Heng Huang, and Hongxia Jin. 2023{\natexlab{a}}.
\newblock \href {http://arxiv.org/abs/2307.08701} {Alpagasus: Training a better alpaca with fewer data}.

\bibitem[{Chen et~al.(2023{\natexlab{b}})Chen, Saifullah, Li, Zhou, and Huang}]{claude2-alpaca}
Lichang Chen, Khalid Saifullah, Ming Li, Tianyi Zhou, and Heng Huang. 2023{\natexlab{b}}.
\newblock Claude2-alpaca: Instruction tuning datasets distilled from claude.
\newblock \url{https://github.com/Lichang-Chen/claude2-alpaca}.

\bibitem[{Chiang and Lee(2023)}]{chiang-lee-2023-large}
Cheng-Han Chiang and Hung-yi Lee. 2023.
\newblock \href {https://doi.org/10.18653/v1/2023.acl-long.870} {Can large language models be an alternative to human evaluations?}
\newblock In \emph{Proceedings of the 61st Annual Meeting of the Association for Computational Linguistics (Volume 1: Long Papers)}, pages 15607--15631, Toronto, Canada. Association for Computational Linguistics.

\bibitem[{Chiang et~al.(2023)Chiang, Li, Lin, Sheng, Wu, Zhang, Zheng, Zhuang, Zhuang, Gonzalez, Stoica, and Xing}]{vicuna2023}
Wei-Lin Chiang, Zhuohan Li, Zi~Lin, Ying Sheng, Zhanghao Wu, Hao Zhang, Lianmin Zheng, Siyuan Zhuang, Yonghao Zhuang, Joseph~E. Gonzalez, Ion Stoica, and Eric~P. Xing. 2023.
\newblock \href {https://lmsys.org/blog/2023-03-30-vicuna/} {Vicuna: An open-source chatbot impressing gpt-4 with 90\%* chatgpt quality}.

\bibitem[{Chung et~al.(2022)Chung, Hou, Longpre, Zoph, Tay, Fedus, Li, Wang, Dehghani, Brahma, Webson, Gu, Dai, Suzgun, Chen, Chowdhery, Valter, Narang, Mishra, Yu, Zhao, Huang, Dai, Yu, Petrov, hsin Chi, Dean, Devlin, Roberts, Zhou, Le, and Wei}]{Chung2022ScalingIL}
Hyung~Won Chung, Le~Hou, S.~Longpre, Barret Zoph, Yi~Tay, William Fedus, Eric Li, Xuezhi Wang, Mostafa Dehghani, Siddhartha Brahma, Albert Webson, Shixiang~Shane Gu, Zhuyun Dai, Mirac Suzgun, Xinyun Chen, Aakanksha Chowdhery, Dasha Valter, Sharan Narang, Gaurav Mishra, Adams~Wei Yu, Vincent Zhao, Yanping Huang, Andrew~M. Dai, Hongkun Yu, Slav Petrov, Ed~Huai hsin Chi, Jeff Dean, Jacob Devlin, Adam Roberts, Denny Zhou, Quoc~V. Le, and Jason Wei. 2022.
\newblock \href {https://api.semanticscholar.org/CorpusID:253018554} {Scaling instruction-finetuned language models}.
\newblock \emph{ArXiv}, abs/2210.11416.

\bibitem[{Clark et~al.(2018)Clark, Cowhey, Etzioni, Khot, Sabharwal, Schoenick, and Tafjord}]{clark2018think}
Peter Clark, Isaac Cowhey, Oren Etzioni, Tushar Khot, Ashish Sabharwal, Carissa Schoenick, and Oyvind Tafjord. 2018.
\newblock \href {http://arxiv.org/abs/1803.05457} {Think you have solved question answering? try arc, the ai2 reasoning challenge}.

\bibitem[{Conover et~al.(2023)Conover, Hayes, Mathur, Xie, Wan, Shah, Ghodsi, Wendell, Zaharia, and Xin}]{DatabricksBlog2023DollyV2}
Mike Conover, Matt Hayes, Ankit Mathur, Jianwei Xie, Jun Wan, Sam Shah, Ali Ghodsi, Patrick Wendell, Matei Zaharia, and Reynold Xin. 2023.
\newblock \href {https://www.databricks.com/blog/2023/04/12/dolly-first-open-commercially-viable-instruction-tuned-llm} {Free dolly: Introducing the world's first truly open instruction-tuned llm}.

\bibitem[{Dao et~al.(2022)Dao, Fu, Ermon, Rudra, and Ré}]{dao2022flashattention}
Tri Dao, Daniel~Y. Fu, Stefano Ermon, Atri Rudra, and Christopher Ré. 2022.
\newblock \href {http://arxiv.org/abs/2205.14135} {Flashattention: Fast and memory-efficient exact attention with io-awareness}.

\bibitem[{Dettmers et~al.(2023)Dettmers, Pagnoni, Holtzman, and Zettlemoyer}]{dettmers2023qlora}
Tim Dettmers, Artidoro Pagnoni, Ari Holtzman, and Luke Zettlemoyer. 2023.
\newblock \href {http://arxiv.org/abs/2305.14314} {Qlora: Efficient finetuning of quantized llms}.

\bibitem[{Ding et~al.(2023)Ding, Chen, Xu, Qin, Hu, Liu, Sun, and Zhou}]{ding2023enhancing}
Ning Ding, Yulin Chen, Bokai Xu, Yujia Qin, Shengding Hu, Zhiyuan Liu, Maosong Sun, and Bowen Zhou. 2023.
\newblock \href {https://doi.org/10.18653/v1/2023.emnlp-main.183} {Enhancing chat language models by scaling high-quality instructional conversations}.
\newblock In \emph{Proceedings of the 2023 Conference on Empirical Methods in Natural Language Processing}, pages 3029--3051, Singapore. Association for Computational Linguistics.

\bibitem[{Du et~al.(2022)Du, Qian, Liu, Ding, Qiu, Yang, and Tang}]{du-etal-2022-glm}
Zhengxiao Du, Yujie Qian, Xiao Liu, Ming Ding, Jiezhong Qiu, Zhilin Yang, and Jie Tang. 2022.
\newblock \href {https://doi.org/10.18653/v1/2022.acl-long.26} {{GLM}: General language model pretraining with autoregressive blank infilling}.
\newblock In \emph{Proceedings of the 60th Annual Meeting of the Association for Computational Linguistics (Volume 1: Long Papers)}, pages 320--335, Dublin, Ireland. Association for Computational Linguistics.

\bibitem[{Dubois et~al.(2023)Dubois, Li, Taori, Zhang, Gulrajani, Ba, Guestrin, Liang, and Hashimoto}]{dubois2023alpacafarm}
Yann Dubois, Xuechen Li, Rohan Taori, Tianyi Zhang, Ishaan Gulrajani, Jimmy Ba, Carlos Guestrin, Percy Liang, and Tatsunori~B. Hashimoto. 2023.
\newblock \href {http://arxiv.org/abs/2305.14387} {Alpacafarm: A simulation framework for methods that learn from human feedback}.

\bibitem[{Gao et~al.(2021)Gao, Tow, Biderman, Black, DiPofi, Foster, Golding, Hsu, McDonell, Muennighoff, Phang, Reynolds, Tang, Thite, Wang, Wang, and Zou}]{eval-harness}
Leo Gao, Jonathan Tow, Stella Biderman, Sid Black, Anthony DiPofi, Charles Foster, Laurence Golding, Jeffrey Hsu, Kyle McDonell, Niklas Muennighoff, Jason Phang, Laria Reynolds, Eric Tang, Anish Thite, Ben Wang, Kevin Wang, and Andy Zou. 2021.
\newblock \href {https://doi.org/10.5281/zenodo.5371628} {A framework for few-shot language model evaluation}.

\bibitem[{Hendrycks et~al.(2021)Hendrycks, Burns, Basart, Zou, Mazeika, Song, and Steinhardt}]{hendrycks2021measuring}
Dan Hendrycks, Collin Burns, Steven Basart, Andy Zou, Mantas Mazeika, Dawn Song, and Jacob Steinhardt. 2021.
\newblock \href {https://openreview.net/forum?id=d7KBjmI3GmQ} {Measuring massive multitask language understanding}.
\newblock In \emph{International Conference on Learning Representations}.

\bibitem[{Huang et~al.(2023)Huang, Gu, Hou, Wu, Wang, Yu, and Han}]{self-improve}
Jiaxin Huang, Shixiang Gu, Le~Hou, Yuexin Wu, Xuezhi Wang, Hongkun Yu, and Jiawei Han. 2023.
\newblock \href {https://doi.org/10.18653/v1/2023.emnlp-main.67} {Large language models can self-improve}.
\newblock In \emph{Proceedings of the 2023 Conference on Empirical Methods in Natural Language Processing}, pages 1051--1068, Singapore. Association for Computational Linguistics.

\bibitem[{Khashabi et~al.(2020)Khashabi, Min, Khot, Sabharwal, Tafjord, Clark, and Hajishirzi}]{khashabi-etal-2020-unifiedqa}
Daniel Khashabi, Sewon Min, Tushar Khot, Ashish Sabharwal, Oyvind Tafjord, Peter Clark, and Hannaneh Hajishirzi. 2020.
\newblock \href {https://doi.org/10.18653/v1/2020.findings-emnlp.171} {{UNIFIEDQA}: Crossing format boundaries with a single {QA} system}.
\newblock In \emph{Findings of the Association for Computational Linguistics: EMNLP 2020}, pages 1896--1907, Online. Association for Computational Linguistics.

\bibitem[{Kingma and Ba(2017)}]{kingma2017adam}
Diederik~P. Kingma and Jimmy Ba. 2017.
\newblock \href {http://arxiv.org/abs/1412.6980} {Adam: A method for stochastic optimization}.

\bibitem[{Ko et~al.(2020)Ko, Lee, Kim, Kim, and Kang}]{ko-etal-2020-look}
Miyoung Ko, Jinhyuk Lee, Hyunjae Kim, Gangwoo Kim, and Jaewoo Kang. 2020.
\newblock \href {https://doi.org/10.18653/v1/2020.emnlp-main.84} {Look at the first sentence: Position bias in question answering}.
\newblock In \emph{Proceedings of the 2020 Conference on Empirical Methods in Natural Language Processing (EMNLP)}, pages 1109--1121, Online. Association for Computational Linguistics.

\bibitem[{Lee et~al.(2023)Lee, Phatale, Mansoor, Mesnard, Ferret, Lu, Bishop, Hall, Carbune, Rastogi, and Prakash}]{lee2023rlaif}
Harrison Lee, Samrat Phatale, Hassan Mansoor, Thomas Mesnard, Johan Ferret, Kellie Lu, Colton Bishop, Ethan Hall, Victor Carbune, Abhinav Rastogi, and Sushant Prakash. 2023.
\newblock \href {http://arxiv.org/abs/2309.00267} {Rlaif: Scaling reinforcement learning from human feedback with ai feedback}.

\bibitem[{Li et~al.(2023{\natexlab{a}})Li, Sun, Yuan, Fan, Zhao, and Liu}]{li2023generative}
Junlong Li, Shichao Sun, Weizhe Yuan, Run-Ze Fan, Hai Zhao, and Pengfei Liu. 2023{\natexlab{a}}.
\newblock \href {http://arxiv.org/abs/2310.05470} {Generative judge for evaluating alignment}.

\bibitem[{Li et~al.(2024{\natexlab{a}})Li, Chen, Chen, and Zhou}]{Li2024CanLS}
Ming Li, Jiuhai Chen, Lichang Chen, and Tianyi Zhou. 2024{\natexlab{a}}.
\newblock \href {https://api.semanticscholar.org/CorpusID:267740270} {Can llms speak for diverse people? tuning llms via debate to generate controllable controversial statements}.
\newblock \emph{ArXiv}, abs/2402.10614.

\bibitem[{Li et~al.(2023{\natexlab{b}})Li, Chen, Chen, He, and Zhou}]{li2023reflectiontuning}
Ming Li, Lichang Chen, Jiuhai Chen, Shwai He, and Tianyi Zhou. 2023{\natexlab{b}}.
\newblock \href {https://openreview.net/forum?id=xaqoZZqkPU} {Reflection-tuning: Recycling data for better instruction-tuning}.
\newblock In \emph{NeurIPS 2023 Workshop on Instruction Tuning and Instruction Following}.

\bibitem[{Li et~al.(2024{\natexlab{b}})Li, Zhang, He, Li, Zhao, Wang, Cheng, and Zhou}]{Li2024SuperfilteringWD}
Ming Li, Yong Zhang, Shwai He, Zhitao Li, Hongyu Zhao, Jianzong Wang, Ning Cheng, and Tianyi Zhou. 2024{\natexlab{b}}.
\newblock \href {https://api.semanticscholar.org/CorpusID:267365346} {Superfiltering: Weak-to-strong data filtering for fast instruction-tuning}.
\newblock \emph{ArXiv}, abs/2402.00530.

\bibitem[{Li et~al.(2023{\natexlab{c}})Li, Zhang, Li, Chen, Chen, Cheng, Wang, Zhou, and Xiao}]{li2023quantity}
Ming Li, Yong Zhang, Zhitao Li, Jiuhai Chen, Lichang Chen, Ning Cheng, Jianzong Wang, Tianyi Zhou, and Jing Xiao. 2023{\natexlab{c}}.
\newblock \href {https://api.semanticscholar.org/CorpusID:261076515} {From quantity to quality: Boosting llm performance with self-guided data selection for instruction tuning}.
\newblock \emph{ArXiv}, abs/2308.12032.

\bibitem[{Li et~al.(2023{\natexlab{d}})Li, Yu, Zhou, Schick, Zettlemoyer, Levy, Weston, and Lewis}]{li2023self}
Xian Li, Ping Yu, Chunting Zhou, Timo Schick, Luke Zettlemoyer, Omer Levy, Jason Weston, and Mike Lewis. 2023{\natexlab{d}}.
\newblock \href {http://arxiv.org/abs/2308.06259} {Self-alignment with instruction backtranslation}.

\bibitem[{Li et~al.(2023{\natexlab{e}})Li, Zhang, Dubois, Taori, Gulrajani, Guestrin, Liang, and Hashimoto}]{alpaca_eval}
Xuechen Li, Tianyi Zhang, Yann Dubois, Rohan Taori, Ishaan Gulrajani, Carlos Guestrin, Percy Liang, and Tatsunori~B. Hashimoto. 2023{\natexlab{e}}.
\newblock Alpacaeval: An automatic evaluator of instruction-following models.
\newblock \url{https://github.com/tatsu-lab/alpaca_eval}.

\bibitem[{Lin et~al.(2022)Lin, Hilton, and Evans}]{lin-etal-2022-truthfulqa}
Stephanie Lin, Jacob Hilton, and Owain Evans. 2022.
\newblock \href {https://doi.org/10.18653/v1/2022.acl-long.229} {{T}ruthful{QA}: Measuring how models mimic human falsehoods}.
\newblock In \emph{Proceedings of the 60th Annual Meeting of the Association for Computational Linguistics (Volume 1: Long Papers)}, pages 3214--3252, Dublin, Ireland. Association for Computational Linguistics.

\bibitem[{Liu et~al.(2023{\natexlab{a}})Liu, Wang, Yao, Chen, Song, Cho, Yacoob, and Yu}]{liu2023mmc}
Fuxiao Liu, Xiaoyang Wang, Wenlin Yao, Jianshu Chen, Kaiqiang Song, Sangwoo Cho, Yaser Yacoob, and Dong Yu. 2023{\natexlab{a}}.
\newblock \href {http://arxiv.org/abs/2311.10774} {Mmc: Advancing multimodal chart understanding with large-scale instruction tuning}.

\bibitem[{Liu et~al.(2023{\natexlab{b}})Liu, Zeng, He, Jiang, and He}]{liu2023makes}
Wei Liu, Weihao Zeng, Keqing He, Yong Jiang, and Junxian He. 2023{\natexlab{b}}.
\newblock \href {http://arxiv.org/abs/2312.15685} {What makes good data for alignment? a comprehensive study of automatic data selection in instruction tuning}.

\bibitem[{Liu et~al.(2023{\natexlab{c}})Liu, Iter, Xu, Wang, Xu, and Zhu}]{liu2023geval}
Yang Liu, Dan Iter, Yichong Xu, Shuohang Wang, Ruochen Xu, and Chenguang Zhu. 2023{\natexlab{c}}.
\newblock \href {http://arxiv.org/abs/2303.16634} {G-eval: Nlg evaluation using gpt-4 with better human alignment}.

\bibitem[{Longpre et~al.(2023)Longpre, Hou, Vu, Webson, Chung, Tay, Zhou, Le, Zoph, Wei, and Roberts}]{Longpre2023TheFC}
S.~Longpre, Le~Hou, Tu~Vu, Albert Webson, Hyung~Won Chung, Yi~Tay, Denny Zhou, Quoc~V. Le, Barret Zoph, Jason Wei, and Adam Roberts. 2023.
\newblock \href {https://api.semanticscholar.org/CorpusID:256415991} {The flan collection: Designing data and methods for effective instruction tuning}.
\newblock \emph{ArXiv}, abs/2301.13688.

\bibitem[{Mishra et~al.(2022)Mishra, Khashabi, Baral, and Hajishirzi}]{mishra2021cross}
Swaroop Mishra, Daniel Khashabi, Chitta Baral, and Hannaneh Hajishirzi. 2022.
\newblock \href {https://doi.org/10.18653/v1/2022.acl-long.244} {Cross-task generalization via natural language crowdsourcing instructions}.
\newblock In \emph{Proceedings of the 60th Annual Meeting of the Association for Computational Linguistics (Volume 1: Long Papers)}, pages 3470--3487, Dublin, Ireland. Association for Computational Linguistics.

\bibitem[{Mitra et~al.(2023)Mitra, Corro, Mahajan, Codas, Simoes, Agrawal, Chen, Razdaibiedina, Jones, Aggarwal, Palangi, Zheng, Rosset, Khanpour, and Awadallah}]{mitra2023orca}
Arindam Mitra, Luciano~Del Corro, Shweti Mahajan, Andres Codas, Clarisse Simoes, Sahaj Agrawal, Xuxi Chen, Anastasia Razdaibiedina, Erik Jones, Kriti Aggarwal, Hamid Palangi, Guoqing Zheng, Corby Rosset, Hamed Khanpour, and Ahmed Awadallah. 2023.
\newblock \href {http://arxiv.org/abs/2311.11045} {Orca 2: Teaching small language models how to reason}.

\bibitem[{OpenAI(2023)}]{openai2023gpt4}
OpenAI. 2023.
\newblock \href {http://arxiv.org/abs/2303.08774} {Gpt-4 technical report}.

\bibitem[{Ouyang et~al.(2022)Ouyang, Wu, Jiang, Almeida, Wainwright, Mishkin, Zhang, Agarwal, Slama, Ray, Schulman, Hilton, Kelton, Miller, Simens, Askell, Welinder, Christiano, Leike, and Lowe}]{NEURIPS2022_b1efde53}
Long Ouyang, Jeffrey Wu, Xu~Jiang, Diogo Almeida, Carroll Wainwright, Pamela Mishkin, Chong Zhang, Sandhini Agarwal, Katarina Slama, Alex Ray, John Schulman, Jacob Hilton, Fraser Kelton, Luke Miller, Maddie Simens, Amanda Askell, Peter Welinder, Paul~F Christiano, Jan Leike, and Ryan Lowe. 2022.
\newblock \href {https://proceedings.neurips.cc/paper_files/paper/2022/file/b1efde53be364a73914f58805a001731-Paper-Conference.pdf} {Training language models to follow instructions with human feedback}.
\newblock In \emph{Advances in Neural Information Processing Systems}, volume~35, pages 27730--27744. Curran Associates, Inc.

\bibitem[{Pan et~al.(2023)Pan, Saxon, Xu, Nathani, Wang, and Wang}]{pan2023automatically}
Liangming Pan, Michael Saxon, Wenda Xu, Deepak Nathani, Xinyi Wang, and William~Yang Wang. 2023.
\newblock \href {http://arxiv.org/abs/2308.03188} {Automatically correcting large language models: Surveying the landscape of diverse self-correction strategies}.

\bibitem[{Peng et~al.(2023)Peng, Li, He, Galley, and Gao}]{peng2023instruction}
Baolin Peng, Chunyuan Li, Pengcheng He, Michel Galley, and Jianfeng Gao. 2023.
\newblock \href {http://arxiv.org/abs/2304.03277} {Instruction tuning with gpt-4}.

\bibitem[{Reimers and Gurevych(2019)}]{reimers-gurevych-2019-sentence}
Nils Reimers and Iryna Gurevych. 2019.
\newblock \href {https://doi.org/10.18653/v1/D19-1410} {Sentence-{BERT}: Sentence embeddings using {S}iamese {BERT}-networks}.
\newblock In \emph{Proceedings of the 2019 Conference on Empirical Methods in Natural Language Processing and the 9th International Joint Conference on Natural Language Processing (EMNLP-IJCNLP)}, pages 3982--3992, Hong Kong, China. Association for Computational Linguistics.

\bibitem[{Taori et~al.(2023)Taori, Gulrajani, Zhang, Dubois, Li, Guestrin, Liang, and Hashimoto}]{alpaca}
Rohan Taori, Ishaan Gulrajani, Tianyi Zhang, Yann Dubois, Xuechen Li, Carlos Guestrin, Percy Liang, and Tatsunori~B. Hashimoto. 2023.
\newblock Stanford alpaca: An instruction-following llama model.
\newblock \url{https://github.com/tatsu-lab/stanford_alpaca}.

\bibitem[{Team(2023)}]{xwin-lm}
Xwin-LM Team. 2023.
\newblock \href {https://github.com/Xwin-LM/Xwin-LM} {Xwin-lm}.

\bibitem[{Touvron et~al.(2023)Touvron, Martin, Stone, Albert, Almahairi, Babaei, Bashlykov, Batra, Bhargava, Bhosale, Bikel, Blecher, Ferrer, Chen, Cucurull, Esiobu, Fernandes, Fu, Fu, Fuller, Gao, Goswami, Goyal, Hartshorn, Hosseini, Hou, Inan, Kardas, Kerkez, Khabsa, Kloumann, Korenev, Koura, Lachaux, Lavril, Lee, Liskovich, Lu, Mao, Martinet, Mihaylov, Mishra, Molybog, Nie, Poulton, Reizenstein, Rungta, Saladi, Schelten, Silva, Smith, Subramanian, Tan, Tang, Taylor, Williams, Kuan, Xu, Yan, Zarov, Zhang, Fan, Kambadur, Narang, Rodriguez, Stojnic, Edunov, and Scialom}]{touvron2023llama2}
Hugo Touvron, Louis Martin, Kevin Stone, Peter Albert, Amjad Almahairi, Yasmine Babaei, Nikolay Bashlykov, Soumya Batra, Prajjwal Bhargava, Shruti Bhosale, Dan Bikel, Lukas Blecher, Cristian~Canton Ferrer, Moya Chen, Guillem Cucurull, David Esiobu, Jude Fernandes, Jeremy Fu, Wenyin Fu, Brian Fuller, Cynthia Gao, Vedanuj Goswami, Naman Goyal, Anthony Hartshorn, Saghar Hosseini, Rui Hou, Hakan Inan, Marcin Kardas, Viktor Kerkez, Madian Khabsa, Isabel Kloumann, Artem Korenev, Punit~Singh Koura, Marie-Anne Lachaux, Thibaut Lavril, Jenya Lee, Diana Liskovich, Yinghai Lu, Yuning Mao, Xavier Martinet, Todor Mihaylov, Pushkar Mishra, Igor Molybog, Yixin Nie, Andrew Poulton, Jeremy Reizenstein, Rashi Rungta, Kalyan Saladi, Alan Schelten, Ruan Silva, Eric~Michael Smith, Ranjan Subramanian, Xiaoqing~Ellen Tan, Binh Tang, Ross Taylor, Adina Williams, Jian~Xiang Kuan, Puxin Xu, Zheng Yan, Iliyan Zarov, Yuchen Zhang, Angela Fan, Melanie Kambadur, Sharan Narang, Aurelien Rodriguez, Robert Stojnic, Sergey Edunov, and Thomas
  Scialom. 2023.
\newblock \href {http://arxiv.org/abs/2307.09288} {Llama 2: Open foundation and fine-tuned chat models}.

\bibitem[{Tunstall et~al.(2023)Tunstall, Beeching, Lambert, Rajani, Rasul, Belkada, Huang, von Werra, Fourrier, Habib, Sarrazin, Sanseviero, Rush, and Wolf}]{tunstall2023zephyr}
Lewis Tunstall, Edward Beeching, Nathan Lambert, Nazneen Rajani, Kashif Rasul, Younes Belkada, Shengyi Huang, Leandro von Werra, Clémentine Fourrier, Nathan Habib, Nathan Sarrazin, Omar Sanseviero, Alexander~M. Rush, and Thomas Wolf. 2023.
\newblock \href {http://arxiv.org/abs/2310.16944} {Zephyr: Direct distillation of lm alignment}.

\bibitem[{Vu et~al.(2023)Vu, He, Haffari, and Shareghi}]{vu2023koala}
Thuy-Trang Vu, Xuanli He, Gholamreza Haffari, and Ehsan Shareghi. 2023.
\newblock \href {http://arxiv.org/abs/2303.14770} {Koala: An index for quantifying overlaps with pre-training corpora}.

\bibitem[{Wang et~al.(2023{\natexlab{a}})Wang, Cheng, Zhan, Li, Song, and Liu}]{wang2023openchat}
Guan Wang, Sijie Cheng, Xianyuan Zhan, Xiangang Li, Sen Song, and Yang Liu. 2023{\natexlab{a}}.
\newblock \href {http://arxiv.org/abs/2309.11235} {Openchat: Advancing open-source language models with mixed-quality data}.

\bibitem[{Wang et~al.(2023{\natexlab{b}})Wang, Li, Chen, Zhu, Lin, Cao, Liu, Liu, and Sui}]{wang2023large}
Peiyi Wang, Lei Li, Liang Chen, Dawei Zhu, Binghuai Lin, Yunbo Cao, Qi~Liu, Tianyu Liu, and Zhifang Sui. 2023{\natexlab{b}}.
\newblock \href {http://arxiv.org/abs/2305.17926} {Large language models are not fair evaluators}.

\bibitem[{Wang et~al.(2023{\natexlab{c}})Wang, Yu, Tan, O'Brien, Pasunuru, Dwivedi-Yu, Golovneva, Zettlemoyer, Fazel-Zarandi, and Celikyilmaz}]{wang2023shepherd}
Tianlu Wang, Ping Yu, Xiaoqing~Ellen Tan, Sean O'Brien, Ramakanth Pasunuru, Jane Dwivedi-Yu, Olga Golovneva, Luke Zettlemoyer, Maryam Fazel-Zarandi, and Asli Celikyilmaz. 2023{\natexlab{c}}.
\newblock \href {http://arxiv.org/abs/2308.04592} {Shepherd: A critic for language model generation}.

\bibitem[{Wang et~al.(2023{\natexlab{d}})Wang, Kordi, Mishra, Liu, Smith, Khashabi, and Hajishirzi}]{wang-etal-2023-self-instruct}
Yizhong Wang, Yeganeh Kordi, Swaroop Mishra, Alisa Liu, Noah~A. Smith, Daniel Khashabi, and Hannaneh Hajishirzi. 2023{\natexlab{d}}.
\newblock \href {https://aclanthology.org/2023.acl-long.754} {Self-instruct: Aligning language models with self-generated instructions}.
\newblock In \emph{Proceedings of the 61st Annual Meeting of the Association for Computational Linguistics (Volume 1: Long Papers)}, pages 13484--13508, Toronto, Canada. Association for Computational Linguistics.

\bibitem[{Wang et~al.(2022)Wang, Mishra, Alipoormolabashi, Kordi, Mirzaei, Naik, Ashok, Dhanasekaran, Arunkumar, Stap, Pathak, Karamanolakis, Lai, Purohit, Mondal, Anderson, Kuznia, Doshi, Pal, Patel, Moradshahi, Parmar, Purohit, Varshney, Kaza, Verma, Puri, Karia, Doshi, Sampat, Mishra, Reddy~A, Patro, Dixit, and Shen}]{wang-etal-2022-super}
Yizhong Wang, Swaroop Mishra, Pegah Alipoormolabashi, Yeganeh Kordi, Amirreza Mirzaei, Atharva Naik, Arjun Ashok, Arut~Selvan Dhanasekaran, Anjana Arunkumar, David Stap, Eshaan Pathak, Giannis Karamanolakis, Haizhi Lai, Ishan Purohit, Ishani Mondal, Jacob Anderson, Kirby Kuznia, Krima Doshi, Kuntal~Kumar Pal, Maitreya Patel, Mehrad Moradshahi, Mihir Parmar, Mirali Purohit, Neeraj Varshney, Phani~Rohitha Kaza, Pulkit Verma, Ravsehaj~Singh Puri, Rushang Karia, Savan Doshi, Shailaja~Keyur Sampat, Siddhartha Mishra, Sujan Reddy~A, Sumanta Patro, Tanay Dixit, and Xudong Shen. 2022.
\newblock \href {https://aclanthology.org/2022.emnlp-main.340} {Super-{N}atural{I}nstructions: Generalization via declarative instructions on 1600+ {NLP} tasks}.
\newblock In \emph{Proceedings of the 2022 Conference on Empirical Methods in Natural Language Processing}, pages 5085--5109, Abu Dhabi, United Arab Emirates. Association for Computational Linguistics.

\bibitem[{Wei et~al.(2022)Wei, Bosma, Zhao, Guu, Yu, Lester, Du, Dai, and Le}]{wei2022finetuned}
Jason Wei, Maarten Bosma, Vincent Zhao, Kelvin Guu, Adams~Wei Yu, Brian Lester, Nan Du, Andrew~M. Dai, and Quoc~V Le. 2022.
\newblock \href {https://openreview.net/forum?id=gEZrGCozdqR} {Finetuned language models are zero-shot learners}.
\newblock In \emph{International Conference on Learning Representations}.

\bibitem[{Wei et~al.(2023)Wei, Wang, Schuurmans, Bosma, Ichter, Xia, Chi, Le, and Zhou}]{wei2023chainofthought}
Jason Wei, Xuezhi Wang, Dale Schuurmans, Maarten Bosma, Brian Ichter, Fei Xia, Ed~Chi, Quoc Le, and Denny Zhou. 2023.
\newblock \href {http://arxiv.org/abs/2201.11903} {Chain-of-thought prompting elicits reasoning in large language models}.

\bibitem[{Wu et~al.(2024)Wu, Waheed, Zhang, Abdul-Mageed, and Aji}]{wu2024laminilm}
Minghao Wu, Abdul Waheed, Chiyu Zhang, Muhammad Abdul-Mageed, and Alham~Fikri Aji. 2024.
\newblock \href {http://arxiv.org/abs/2304.14402} {Lamini-lm: A diverse herd of distilled models from large-scale instructions}.

\bibitem[{Xu et~al.(2023{\natexlab{a}})Xu, Sun, Zheng, Geng, Zhao, Feng, Tao, and Jiang}]{xu2023wizardlm}
Can Xu, Qingfeng Sun, Kai Zheng, Xiubo Geng, Pu~Zhao, Jiazhan Feng, Chongyang Tao, and Daxin Jiang. 2023{\natexlab{a}}.
\newblock \href {http://arxiv.org/abs/2304.12244} {Wizardlm: Empowering large language models to follow complex instructions}.

\bibitem[{Xu et~al.(2024)Xu, Li, Tao, Shen, Cheng, Li, Xu, Tao, and Zhou}]{xu2024survey}
Xiaohan Xu, Ming Li, Chongyang Tao, Tao Shen, Reynold Cheng, Jinyang Li, Can Xu, Dacheng Tao, and Tianyi Zhou. 2024.
\newblock \href {http://arxiv.org/abs/2402.13116} {A survey on knowledge distillation of large language models}.

\bibitem[{Xu et~al.(2023{\natexlab{b}})Xu, Yao, Huang, Qi, Wang, Gu, and Sundaresan}]{xu2023rethinking}
Yang Xu, Yongqiang Yao, Yufan Huang, Mengnan Qi, Maoquan Wang, Bin Gu, and Neel Sundaresan. 2023{\natexlab{b}}.
\newblock \href {http://arxiv.org/abs/2312.11508} {Rethinking the instruction quality: Lift is what you need}.

\bibitem[{Yao et~al.(2023)Yao, Yu, Zhao, Shafran, Griffiths, Cao, and Narasimhan}]{yao2023tree}
Shunyu Yao, Dian Yu, Jeffrey Zhao, Izhak Shafran, Thomas~L. Griffiths, Yuan Cao, and Karthik Narasimhan. 2023.
\newblock \href {http://arxiv.org/abs/2305.10601} {Tree of thoughts: Deliberate problem solving with large language models}.

\bibitem[{Ye et~al.(2021)Ye, Lin, and Ren}]{ye-etal-2021-crossfit}
Qinyuan Ye, Bill~Yuchen Lin, and Xiang Ren. 2021.
\newblock \href {https://doi.org/10.18653/v1/2021.emnlp-main.572} {{C}ross{F}it: A few-shot learning challenge for cross-task generalization in {NLP}}.
\newblock In \emph{Proceedings of the 2021 Conference on Empirical Methods in Natural Language Processing}, pages 7163--7189, Online and Punta Cana, Dominican Republic. Association for Computational Linguistics.

\bibitem[{Ye et~al.(2023)Ye, Jo, Kim, Kim, Hwang, and Seo}]{selfee2023}
Seonghyeon Ye, Yongrae Jo, Doyoung Kim, Sungdong Kim, Hyeonbin Hwang, and Minjoon Seo. 2023.
\newblock \href {https://kaistai.github.io/SelFee/} {Selfee: Iterative self-revising llm empowered by self-feedback generation}.
\newblock Blog post.

\bibitem[{Zellers et~al.(2019)Zellers, Holtzman, Bisk, Farhadi, and Choi}]{zellers-etal-2019-hellaswag}
Rowan Zellers, Ari Holtzman, Yonatan Bisk, Ali Farhadi, and Yejin Choi. 2019.
\newblock \href {https://doi.org/10.18653/v1/P19-1472} {{H}ella{S}wag: Can a machine really finish your sentence?}
\newblock In \emph{Proceedings of the 57th Annual Meeting of the Association for Computational Linguistics}, pages 4791--4800, Florence, Italy. Association for Computational Linguistics.

\bibitem[{Zhang et~al.(2023)Zhang, Dong, Li, Zhang, Sun, Wang, Li, Hu, Zhang, Wu, and Wang}]{zhang2023instruction}
Shengyu Zhang, Linfeng Dong, Xiaoya Li, Sen Zhang, Xiaofei Sun, Shuhe Wang, Jiwei Li, Runyi Hu, Tianwei Zhang, Fei Wu, and Guoyin Wang. 2023.
\newblock \href {http://arxiv.org/abs/2308.10792} {Instruction tuning for large language models: A survey}.

\bibitem[{Zheng et~al.(2023)Zheng, Chiang, Sheng, Zhuang, Wu, Zhuang, Lin, Li, Li, Xing, Zhang, Gonzalez, and Stoica}]{zheng2023judging}
Lianmin Zheng, Wei-Lin Chiang, Ying Sheng, Siyuan Zhuang, Zhanghao Wu, Yonghao Zhuang, Zi~Lin, Zhuohan Li, Dacheng Li, Eric.~P Xing, Hao Zhang, Joseph~E. Gonzalez, and Ion Stoica. 2023.
\newblock \href {http://arxiv.org/abs/2306.05685} {Judging llm-as-a-judge with mt-bench and chatbot arena}.

\bibitem[{Zhou et~al.(2023)Zhou, Liu, Xu, Iyer, Sun, Mao, Ma, Efrat, Yu, Yu, Zhang, Ghosh, Lewis, Zettlemoyer, and Levy}]{zhou2023lima}
Chunting Zhou, Pengfei Liu, Puxin Xu, Srini Iyer, Jiao Sun, Yuning Mao, Xuezhe Ma, Avia Efrat, Ping Yu, Lili Yu, Susan Zhang, Gargi Ghosh, Mike Lewis, Luke Zettlemoyer, and Omer Levy. 2023.
\newblock \href {http://arxiv.org/abs/2305.11206} {Lima: Less is more for alignment}.

\end{thebibliography}
